\documentclass[a4paper]{article}

\usepackage[utf8]{inputenc}
\usepackage{import}
\usepackage[margin=1in]{geometry}
\usepackage{setspace}
\usepackage{microtype}
\usepackage{adjustbox}
\usepackage{multirow}
\usepackage{graphicx}
\usepackage[]{subcaption}
\usepackage{algorithm}
\usepackage{algpseudocode}
\usepackage{longtable}
\usepackage{pdflscape}
\usepackage{rotating, tabularx}
\graphicspath{ {./images/} }

\usepackage{amsmath}
\usepackage{amssymb}
\usepackage{bm}
\usepackage{natbib}
\usepackage{amsthm}
\newtheorem{lemma}{Lemma}

\usepackage{booktabs} %
\usepackage[hyphens]{url}
\usepackage[bookmarks=false,hidelinks]{hyperref}

\usepackage{mathtools}
\usepackage{color}

\usepackage{authblk}

\usepackage{tikz}
\usetikzlibrary{matrix,backgrounds}
\usetikzlibrary{decorations.pathreplacing}
\usetikzlibrary{positioning}
\usetikzlibrary{calc}

\tikzstyle{chromosome} = [
	matrix of math nodes, ampersand replacement=\&,
	nodes={draw, anchor=center, minimum width=2.5em},
	nodes in empty cells, minimum height=1.8em
]

\tikzstyle{depotnode} = [shape=circle, draw, fill=white!20!black, text=white]
\tikzstyle{customernode} = [shape=circle, draw, fill=white!40!blue, text=white]

\newcommand{\email}[1]{{\href{mailto:#1}{\nolinkurl{#1}}}}

\usepackage{bm}

\newcommand{\set}[1]{\mathcal{#1}}

\usepackage{xcolor}

\newcommand{\bluenote}[1]{#1}

\usepackage{etoolbox}
\makeatletter
\def\do#1{\@namedef{#1c}{\ensuremath{\mathcal{#1}}}}
\docsvlist{A, B, C, D, E, F, G, H, I, J, K, L, M, N, O, P, Q, R, 
S, T, U, V, W, X, Y, Z}
\makeatother

\usepackage{bbm}

\renewcommand{\bar}[1]{\mkern 1.5mu\overline{\mkern-1.5mu#1\mkern-1.5mu}\mkern 1.5mu}

\newcommand{\tr}{\mathrm{tr}}
\newcommand{\dr}{\mathrm{dr}}    
\newcommand{\EP}{\textsc{Partition}}
\newcommand{\SPLIT}{\textsc{Split}}
\newcommand{\JOIN}{\textsc{Join}}

\algnewcommand\algorithmicforeach{\textbf{for each}}
\algdef{S}[FOR]{ForEach}[1]{\algorithmicforeach\ #1\ \algorithmicdo}

\title{A Hybrid Genetic Algorithm with Type-Aware
Chromosomes for Traveling Salesman Problems with
Drone}

\author{Sasan Mahmoudinazlou}
\affil{Department of Industrial and Management Systems Engineering, University of South Florida, Tampa, FL 33620, U.S.A.\\\email{sasanm@usf.edu}}
\author{Changhyun Kwon\thanks{Corresponding author}}
\affil{Department of Industrial and Systems Engineering, KAIST, Daejeon, 34141, Republic of Korea\\\email{chkwon@kaist.ac.kr}}
\affil{OMELET, Daejeon, 34051, Republic of Korea}

\date{}

\begin{document}

\maketitle

\begin{abstract}
There are emerging transportation problems known as the Traveling Salesman Problem with Drone (TSPD) and the Flying Sidekick Traveling Salesman Problem (FSTSP) that involve using a drone in conjunction with a truck for package delivery. This study presents a hybrid genetic algorithm for solving TSPD and FSTSP by incorporating local search and dynamic programming. Similar algorithms exist in the literature. Our algorithm, however, considers more sophisticated chromosomes and less computationally complex dynamic programming to enable broader exploration by the genetic algorithm and efficient exploitation through dynamic programming and local search. The key contribution of this paper is the discovery of how decision-making processes for solving TSPD and FSTSP should be divided among the layers of genetic algorithm, dynamic programming, and local search. In particular, our genetic algorithm generates the truck and the drone sequences separately and encodes them in a type-aware chromosome, wherein each customer is assigned to either the truck or the drone. We apply local search to each chromosome, which is decoded by dynamic programming for fitness evaluation. Our new algorithm is shown to outperform existing algorithms on most benchmark instances in both quality and time. Our algorithms found the new best solutions for 538 TSPD instances out of 920 and 74 FSTSP instances out of 132.

\end{abstract}

\section{Introduction}

With the growing number of online orders, companies are competing to ship products in a timely manner.
This underscores the need for improved delivery methods.
Traditional methods include using the Traveling Salesman Problem (TSP), in which a vehicle, perhaps a truck, takes multiple orders and delivers them to customers as soon as possible. 
As a result of the truck's low speed and need to maneuver around traffic flows, many large companies such as Amazon and UPS are looking for new delivery methods like drones. 
Unlike traditional trucks, drones do not have to follow the same network, giving them faster delivery times. 
They are also more eco-friendly since they are electrical and less expensive since they do not require any human involvement. 
    
It is important to note that drones also come with some significant limitations.
Unlike trucks, drones do not have the capability to carry heavy or multiple parcels. 
To be able to pick up an item after every delivery, a drone must return to the warehouse. 
Drones are limited in their range due to their battery capacity and cannot reach remote customers. 

Recent efforts have been made to utilize the advantages of both vehicles and include a delivery method in which the truck and the drone collaborate. 
\citet{murray2015flying} describe the combined problem as the Flying Sidekick Traveling Salesman Problem (FSTSP).
The FSTSP extends the traditional Traveling Salesman Problem by introducing a ``sidekick'' that can fly and assist the salesperson in visiting cities. 
The challenge is to find the shortest route that allows both the salesperson (ground vehicle) and sidekick (drone) to visit all cities while considering the drone's limited flying range.
The drone lies on top of the truck and can fly at various points, deliver a parcel to a customer, and land on the truck at another point while the truck can make deliveries at the same time. 
A similar problem is introduced by \citet{agatz2018optimization} as the Traveling Salesman Problem with Drone (TSPD) with slightly simpler assumptions.
The main difference is that the drone can wait for the truck in the TSPD but not in the FSTSP, among other differences in the assumptions.
Section \ref{sec:Problem} provides an in-depth elucidation of the assumptions pertaining to two problems, highlighting their distinct characteristics and points of divergence.
We refer readers to \citet{chung2020optimization} and \citet{macrina2020drone} for other problems that involve both ground vehicles and drones.

A hybrid genetic algorithm (HGA) is presented in this study for solving both TSPD and FSTSP with the aim of minimizing total delivery time. 
While regular genetic algorithms only rely on crossovers and mutations to generate new solutions, hybrid genetic algorithms, also known as memetic algorithms \citep{moscato1989evolution}, usually improve the generated solutions by local search or other heuristics in addition to those conventional techniques.
Hybrid genetic algorithms have been actively developed to solve various vehicle routing problems \citep{ho2008hybrid,vidal2012hybrid}, including the problem of our interest, TSPD/FSTSP \citep{ha2020hybrid}.

Our HGA approach consists of three layers in its algorithmic structure. 
First, a regular genetic algorithm (GA) layer generates incomplete solutions by determining the customer nodes served by the truck and the drone, called truck nodes and drone nodes, respectively. 
Second, a dynamic programming (DP) layer completes the solutions by determining the combined nodes, where the drone departs from the truck or returns to the truck. 
Third, we use various local search \bluenote{neighborhoods} to improve the generated solutions.

The key contribution of this paper is the discovery of how decision-making processes should be divided among these three layers.
Similar approaches exist in the literature, especially, \citet{ha2020hybrid}, whose GA layer is simple while the difficult decision-making step is handled by their DP layer.
We give more informed exploratory roles in the GA layer,  while we use a much simpler DP layer so that the decoding and evaluation for exploitation are faster.
Our GA layer utilizes a novel type-aware chromosome (TAC) encoding to distinguish truck and drone nodes.
We also devise type-aware order crossover operations to support the TAC encoding in the GA layer.
Our unique TAC encoding allows the DP layer to have a time complexity of $O(n^2)$, which is lower than other DP-based approaches used in the literature with the time complexities of $O(n^3)$ and $O(n^4)$ (see Section \ref{DP}).
With the TAC encoding, it is easy to devise various local search \bluenote{neighborhoods} directly on the chromosomes. It is unnecessary to apply additional encoding on the decoded solutions from the DP layer, and it is possible to keep high-resolution information on the solutions in the GA populations.
Furthermore, our GA layer uses two or three subpopulations, depending on whether the drone range is limited, and we propose an escaping strategy for preventing GAs from being trapped in local optima. 
The proposed division of three layers with the TAC encoding brings reduction in the objective function values, savings in the computational time, or both, in most benchmark instances.

The remainder of the paper is organized as follows. 
A brief summary of the literature is given in Section \ref{sec:lit}, and
a formal definition of the problem is given in Section \ref{sec:Problem}. 
The methodology is described in Section \ref{sec:Method}, and
the numerical results from the experiments are given in Section \ref{sec:Results}.
Finally, the conclusion and future directions are discussed in Section \ref{sec:Con}.
In the rest of the paper, we call our method the HGA with TAC, or HGA-TAC, to emphasize the significance of the TAC encoding.

\section{Literature Review} \label{sec:lit}
The number of recent publications that are dedicated to the study of last-mile delivery with truck and drone collaboration is rapidly increasing, and because of this, the majority of the articles that we review are those that are pertinent to our work.
We refer to survey papers, \citet{otto2018optimization}, \citet{macrina2020drone}, \citet{chung2020optimization}, and \citet{li2021ground}, for general overviews on the optimization of problems with various applications of drones. 
Several approaches have been taken in the literature to study the collaboration between trucks and drones. 
Note that the research topics differ regarding the number of vehicles, the basic assumptions and limitations, and the method of solving them. 
As \bluenote{our} study focuses on the collaboration between a single truck and a single drone with only one visit per flight, we review the most relevant papers in this section.

We begin by reviewing the papers that first proposed the problem. 
Our next step is to review publications that employ heuristic and metaheuristic approaches, followed by those that utilize exact approaches. 
Lastly, we introduce two papers that use machine-learning techniques to solve the problem. 

\citet{murray2015flying} introduce the FSTSP, formulate it using mixed-integer linear programming (MILP), and propose a heuristic algorithm for solving it. 
To solve the TSPD, \citet{agatz2018optimization} also propose an MILP formulation, which is computationally tractable for small problems with up to twelve customers.
They provide a heuristic algorithm, called `TSP-ep-all,' that combines neighborhood search and a partitioning algorithm, which begins with an optimal or near-optimal TSP tour. 
Given the initial TSP tour, they use a DP approach to determine which customers should be visited by the drone, while maintaining the order in the given TSP tour.
This procedure is called the \emph{exact partitioning} of customers. 
Then, the TSP-ep-all algorithm generates other TSP tours by perturbing the TSP tour using local search \bluenote{neighborhoods} such as the two-opt swap, the three-opt swap, and the two-point swap, which are popularly used in TSP heuristics. 
For each generated TSP tour, the exact partitioning is used to create a TSPD solution.
The algorithm returns the minimal delivery time TSPD solution as the final solution.
This method has been shown to be efficient and effective in small to medium-sized problems with less than 50 customers, but it shows exponentially increasing computational time due to the extensive local search on the initial TSP tour.
They also created a benchmark TSPD dataset, which we will use to compare our algorithm with other methods.

A Divide-Partition-and-Search heuristic (DPS) was introduced by \citet{bogyrbayeva2023deep} for solving TSPD. 
In the DPS approach, the network is systematically divided into smaller problems, drawing inspiration from the divide-and-conquer heuristic proposed by \citet{poikonen2019branch}. 
Each resulting subproblem is then addressed using the TSP-ep-all algorithm \citep{agatz2018optimization}. 
Specifically, DPS$/g$ partitions all nodes into subgroups, each containing $g$ nodes. 
Subsequently, each subgroup undergoes a partitioning process facilitated by TSP-ep-all. 
It is noteworthy that when $g$ equals the total number of nodes ($N$), DPS$/N$ is equivalent to TSP-ep-all. 
The performance of DPS exhibits improvement with an increase in $g$, albeit at the expense of longer solution times. 
However, DPS$/25$ demonstrates commendable performance, proving to be notably faster than TSP-ep-all, especially for larger instances. 
Consequently, DPS$/25$ is selected as a primary baseline algorithm for comparative analysis in this study.

Various metaheuristic methods have been developed successfully \bluenote{for solving TSPD/FSTSP} in the literature.
Most related to our approach is the Hybrid General Variable Neighborhood Search (HGVNS) of \citet{de2020variable} and the Hybrid Genetic Algorithm (HGA20) of \citet{ha2020hybrid}; not to be confused with our HGA approach, we use the acronym `HGA20' for the method of \citet{ha2020hybrid} throughout the paper. 
HGVNS also starts with an optimal or near-optimal TSP solution and then converts it to an initial solution by creating sub-routes for the drone. 
Finally, HGVNS improves the solution by a variable neighborhood search. 
HGA20 represents each chromosome in the form of a giant tour that corresponds to the TSP tour with the depot removed. 
The \SPLIT{} method, a polynomial time algorithm proposed in \citet{ha2018min}, is employed at each step in order to convert the giant tour chromosome into the FSTSP tour. 
Then, local search \bluenote{neighborhoods} are applied for improvement, and finally, using a restore method, the improved FSTSP tour is converted back into the giant tour chromosome. 
The chromosome encoding in our approach, however, includes the truck sequence as well as the drone sequence. 
Only the locations of drone launches and landings remain \bluenote{undecided}, which are determined by DP in an optimal manner.

Other metaheuristic methods for TSPD/FSTSP include a greedy randomized adaptive search procedure \citep{ha2018min}, a multi-start variable neighborhood search \citep{campuzano2021multi}, and a multi-start tabu search \citep{luo2021multi}.

Numerous studies have attempted to solve TSPD using exact methods.   
A DP approach is developed by \citet{bouman2018dynamic} in order to find an optimal solution to TSPD. 
Their DP algorithm is able to solve the problem with up to 20 customers. 
The branch-and-bound algorithm proposed by \citet{poikonen2019branch} is also capable of solving TSPD up to 20 nodes at optimum. 
Additionally, they devise a divide-and-conquer heuristic based on their branch-and-bound algorithm that delivers high-quality solutions. 
\citet{boccia2021column} introduce a new representation of the FSTSP based on the definition of an extended graph.
They present a new MILP formulation and a branch-and-cut algorithm for solving the problem, which is able to find optimal solutions for problems with up to 20 customers. 
\citet{roberti2021exact} create a new MILP \bluenote{formulation} that is more effective and able to accommodate a variety of side constraints. 
They propose a branch-and-price approach wherein they establish a set partitioning problem and then use the \textit{ng}-route relaxation \citep{baldacci2011new} to formulate the pricing problem. 
Their technique optimally solves TSPD instances for problems with up to 39 customers. 
\citet{yang2023planning} propose a branch-and-price algorithm for solving the TSPD under uncertainty. 
The problem is referred to as the robust drone-truck delivery problem (RDTDP). 
They are able to solve instances of up to 40 customers using their algorithm. 

The use of learning-based methods has been reported in a few papers that address TSPD. 
The K-Means Clustering algorithm is employed by \citet{ferrandez2016optimization} to determine launch locations, and a GA is used to solve a TSP and determine the truck routes.
As part of their objective function, they also aim to minimize the amount of energy consumed as well as the delivery time. 
The study also examines the case of multiple drones per truck. 
\citet{bogyrbayeva2023deep} design and train a deep reinforcement learning algorithm for solving TSPD and examine it on instances with up to 100 nodes in size. 
Aside from the training time, their algorithm is capable of solving instances within a relatively short period of time. 

Most studies have focused either on TSPD or FSTSP, with a few exceptions, such as \citet{de2020variable}, which solves the problem in both settings. 
It is also important to note that the existing algorithms for TSPD are not exempt from the trade-off rule between solution quality and running time. 
This work aims to solve both problems, with a particular focus on the disparities in underlying assumptions, and testing on a range of instances from the literature reveals that our method can provide viable solutions in a reasonable amount of time.

\begin{figure}
    \centering
    \begin{subfigure}[b]{0.45\linewidth}
      \begin{tikzpicture}
          \node[depotnode] (n0) at (0,0) {0};
          \node[customernode] (n1) at (0,2) {1};
          \node[customernode] (n2) at (2,2) {2};				
          \node[customernode] (n3) at (4,1.5) {3};				
          \node[customernode] (n4) at (3,-0.5) {4};
          \node[customernode] (n5) at (1.4,-1.5) {5};		
          \draw[->, thick] (n0) -- node[left] {7} (n1);	
          \draw[->, thick] (n1) -- node[above] {6} (n2);		
          \draw[->, thick] (n2) -- node[above] {8} (n3);
          \draw[->, thick] (n3) -- node[right] {6} (n4);		
          \draw[->, thick] (n4) -- node[below] {7} (n5);		
          \draw[->, thick] (n5) -- node[below] {5} (n0);		
  
          \matrix [draw, below left, right = of n4, 
                  column 1/.style={anchor=base},
              column 2/.style={anchor=base west}
          ] 
          {
              \node [depotnode] {}; 			& \node[] {\tiny Depot}; \\
              \node [customernode] {}; 		& \node[] {\tiny Customer}; \\
              \node [] {$\longrightarrow$};   & \node[] {\tiny Truck arc}; \\
              \node [] {$\dashrightarrow$};   & \node[] {\tiny Drone arc}; \\
          };
      \end{tikzpicture}    
      \caption{The makespan obtained by TSP is 39}
      \label{fig:TSP}
    \end{subfigure}
    \begin{subfigure}[b]{0.45\linewidth}
      \begin{tikzpicture}
          \node[depotnode] (n0) at (0,0) {0};
          \node[customernode] (n1) at (0,2) {1};
          \node[customernode] (n2) at (2,2) {2};				
          \node[customernode] (n3) at (4,1.5) {3};				
          \node[customernode] (n4) at (3,-0.5) {4};
          \node[customernode] (n5) at (1.4,-1.5) {5};		
  
          \draw[->, thick] (n0) -- node[left] {10} (n2);	
          \draw[->, thick] (n2) -- node[right] {6} (n4);
          \draw[->, thick] (n4) -- node[above] {8} (n0);
  
          \draw[->, thick, dashed] (n0) -- node[left] {3.5} (n1);
          \draw[->, thick, dashed] (n1) -- node[above] {3} (n2);
          \draw[->, thick, dashed] (n2) -- node[above] {4} (n3);
          \draw[->, thick, dashed] (n3) -- node[right] {3} (n4);
          \draw[->, thick, dashed] (n4) -- node[below] {3.5} (n5);
          \draw[->, thick, dashed] (n5) -- node[left] {2.5} (n0);
  
          \matrix [draw, below left, right = of n4, 
                  column 1/.style={anchor=base},
              column 2/.style={anchor=base west}
          ] 
          {
              \node [depotnode] {}; 			& \node[] {\tiny Depot}; \\
              \node [customernode] {}; 		& \node[] {\tiny Customer}; \\
              \node [] {$\longrightarrow$};   & \node[] {\tiny Truck arc}; \\
              \node [] {$\dashrightarrow$};   & \node[] {\tiny Drone arc}; \\
          };
      \end{tikzpicture}      
      \caption{The makespan obtained by TSPD is 25}
      \label{fig:TSPD}
    \end{subfigure}
    \caption{A network with one depot and 5 customers solved by TSP and TSPD.}
  \end{figure}
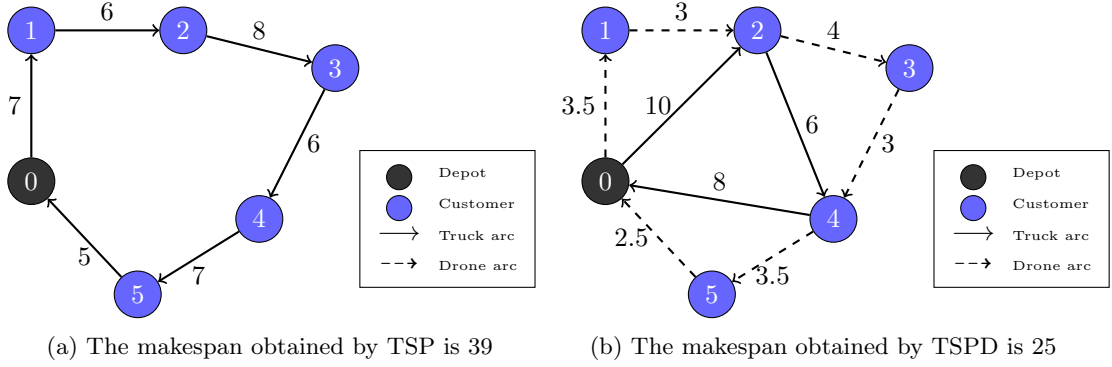
  
\section{Problem Statement} \label{sec:Problem}
In this section, we introduce TSPD and FSTSP formally and describe their associated assumptions.
As explained before, the TSPD/FSTSP system involves coordinating a truck with a drone to deliver goods with the objective of minimizing the total delivery time, also known as the \emph{makespan} in the operations research literature.
For the rest of the paper, we will use the term makespan as our objective function.
As the drone resides on the truck's roof, it can be launched to deliver a package to a customer and then return to the top of the truck. 
The TSPD can be represented on a graph $G(\set{V},\set{E})$ where $\set{V}=\{0, 1, ..., n\}$ is the set of nodes, $n$ is the number of customer nodes, and $\set{E}$ is the set of edges of the graph which is assumed to be complete.
We let $N = |\set{V}| = n+1$.
Vehicles start their routes at the depot node $0$ and must return to it after completing their deliveries; for returning, the depot is represented as $0'$. 
The remaining nodes refer to the locations of customers. 
The travel time between the nodes may vary in accordance with the type of vehicle and the assumptions made about the problem. 
The ratio of drone speed to truck speed is represented by $\alpha$. 
For instance, if $\alpha=2$ and the Euclidean distance is used for both vehicles, the travel time for a drone between two geographical locations will be half of the time required for a truck to travel between the same nodes.

Before delving into the specifics, let us consider an illustrative example of a simple network solved using TSPD and the amount of time saved over regular TSP. 
Figure \ref{fig:TSP} depicts the optimal TSP solution for a network with one depot and five customers, whereas Figure \ref{fig:TSPD} illustrates the same network utilizing a truck and a drone in collaboration.
Numbers on the arcs indicate traversal times for corresponding vehicles with $\alpha=2$. 
The set of deliveries made by the cooperation of two vehicles between each launch and land is named an \emph{operation} by \citet{agatz2018optimization} and \emph{sortie} by \citet{murray2015flying}. 
Each operation is defined as $\langle i,j,k \rangle$, where $i$ is the position where the drone launches, $j$ is the location where the customer is served, and $k$ is the position where the drone lands on the truck. 
In the first operation $\langle 0,1,2 \rangle$, for instance, the drone takes off from the depot, makes a delivery to the customer at node 1, and lands on the truck at node 2. 
The minimum time for each operation can be calculated as $\max\{ \tau_{i \rightarrow k}^{\tr}, \tau_{i,j}^{\dr} + \tau_{j,k}^{\dr}\}$ where $\tau_{i \rightarrow k}^{\tr}$ is truck's travel time from $i$ to $k$ traversing all the nodes in between if any, and $\tau_{i,j}^{\dr}$ is drone's travel time between two nodes. 
In this simplified instance, TSPD shows a 36 percent improvement over TSP, which indicates that drones can serve a valuable and beneficial function for last-mile delivery. 

We first state the common assumptions of the TSPD and FSTSP, and then problem-specific assumptions.

\subsection{Common Assumptions}

Both TSPD and FSTSP make the following assumptions:
\begin{enumerate}
    \item Both vehicles start their tour at the depot and must return there at the end of their tour after completing all deliveries.
    \item In each operation, the drone is only permitted to visit one customer before returning to either the truck or depot. 
        While the drone is in flight, the truck can make multiple deliveries.
    \item Drone launches only at customer locations or depot $0$ and lands only at customer locations or depot $0'$. 
    \item Each customer is visited only once by either the drone or the truck.
        While customers are served by the truck, the drone might be onboard with the truck or in flight. 
\end{enumerate}

\subsection{Problem-Specific Assumptions}
    All of the above general assumptions hold true for both TSPD and FSTSP, whereas the specific assumptions for each problem are listed below.
\begin{enumerate}
    \item \emph{Launch and landing nodes}: In both cases, the drone can be relaunched from the same spot on the vehicle where it landed.
        However, in FSTSP, drones cannot land on the same node from where they are launched, but in TSPD, this is permitted.
    \item \emph{Setup times}: TSPD disregards pickup time, delivery time, and recharging (battery changing) time, whereas FSTSP includes preparation time denoted $s_L$ before a launch for changing the battery and loading the cargo, and retrieval time when landing on the truck denoted $s_R$.
    \item \emph{Drone eligible nodes}: In TSPD, all customers can be visited and serviced by either vehicle; however, in FSTSP, some customers cannot be visited by drone for a variety of reasons. 
        It could be due to a heavy package that cannot be carried by a drone, the necessity for a signature, or an impractical landing site.
    \item \emph{Flying range}: The drone has a limited flight time due to its limited battery capacity. 
        \citet{agatz2018optimization} solves the problem with the assumption of limited and unlimited flying range for the drone.
        In TSPD, the flying range constraint just contains the time that a drone flies between nodes, whereas in FSTSP, it also includes the time that the drone must remain in constant flight if waiting for the truck at the rendezvous point, as it is not permitted to land and wait for the truck. 
        For this reason, each FSTSP operation must satisfy the drone's flying endurance constraint for the truck as well. 
        For clarity, let $e$ represent the drone's endurance. 
        For operation $\langle i,j,k \rangle$, the relevant TSPD constraint for the drone solely is:
        \[\tau_{i,j}^{\dr} + \tau_{j,k}^{\dr} \leq e.\]

        While in the case of FSTSP, the constraint is: 
        \[ \max \{\tau_{i,j}^{\dr} + \tau_{j,k}^{\dr} + s_R , \tau_{i\rightarrow k}^{\tr} + s_T  \}\leq e.\]
        where $\tau_{i\rightarrow k}^{\tr}$ denotes the time required for the truck to travel from node $i$ to $k$ while fulfilling deliveries en route, and $s_T$ represents the time needed for the truck to recover the drone and potentially prepare it for subsequent launches.
        In precise terms, when the drone is solely recovered without relaunching at the same location, the service time $s_T$ is equivalent to the recovery time $s_R$. 
        Conversely, if the truck chooses to relaunch the drone at the same location, the preparation time for the drone's subsequent launch ($s_L$) is added to the service time, resulting in $s_T = s_R + s_L$.
        There is no such constraint when the launch node is the depot.

\end{enumerate}

Despite the fact that these problem names have been used interchangeably in the literature, or some studies may include a combination of these assumptions, we tried to categorize the assumptions based on the first two papers that introduce the problem. 
In this study, we address both of these problems in our algorithm, solve instances from both natures, and compare our results with the best current methods.
These differences in the problem assumptions are handled mainly in the DP layer.

\section{A Hybrid Genetic Algorithm with Type-Aware Chromosomes} \label{sec:Method}
In this section, we describe the structure of our GA, provide the local search \bluenote{neighborhoods}, expound our DP technique, and introduce our strategy for escaping local optima. 

The algorithm we propose is a HGA similar to \citet{vidal2012hybrid} in terms of its basic construction.
We search for both feasible and infeasible solutions stored in two \bluenote{or three} subpopulations. 
Additionally, we take advantage of the contribution each individual provides to the diversity of the gene pool rather than only considering the cost of the solution. 
\citet{ha2020hybrid} also employ this structure for solving FSTSP but \bluenote{use} a completely different representation of solutions and fitness evaluation technique from ours, as it will be explained in Section \ref{DP}.
    
Algorithm \ref{alg:GA} displays our HGA's general structure.
The algorithm commences by initializing the subpopulations, as elucidated in Section \ref{init-pop} (Lines \ref{GAline:init1}-\ref{GAline:init2}).
The feasible, Type 1 infeasible, and Type 2 infeasible subpopulations are denoted by $\Omega_\textsc{f}, \Omega_\textsc{inf}^1$, and $\Omega_\textsc{inf}^2$. 
\bluenote{Individuals with the characteristic of more than one node being visited by the drone in a single flight are placed into the Type 1 infeasible subpopulation. 
Solutions exceeding the drone's range limit are classified into the Type 2 infeasible subpopulation. }
It should be noted that if the range of the drone is unlimited, the use of $\Omega_\textsc{inf}^2$ will be irrelevant, and we would only consider $\Omega_\textsc{f}$ and $\Omega_\textsc{inf}^1$. 

Once the initial population is generated, the iterative process continues until a convergence criterion is met, where convergence is defined as no observed improvement for a predefined number of iterations, denoted as $It_{\textsc{NI}}$. 
At each iteration, two parents are selected from the entire population (Line \ref{GAline:parentselect}), subjected to a crossover operation (Line \ref{GAline:crossover}), resulting in the generation of a new offspring.
This offspring, with a probability $P_\textsc{m}$, undergoes a mutation operation (Line \ref{GAline:mutation}).
Explicit details regarding parent selection, crossover, and mutation are expounded upon in Section \ref{parent}.

\bluenote{After generation}, the offspring undergoes an evaluation to determine its feasibility and corresponding fitness (Line \ref{GAline:feasible}).
If the offspring is deemed infeasible, the fitness incorporates penalties. 
The evaluation process intricacies are outlined in Section \ref{DP}.
In the event of infeasibility, the algorithm makes the solution feasible with probability $P_{\textsc{REPAIR}}$ (Lines \ref{GAline:repair1}, \ref{GAline:repair2}).
The repair procedure is straightforward: when a violation occurs with respect to any of the drone constraints, the associated node \bluenote{changed to be a} truck node. 
Specifically, if the solution entails consecutive drone nodes within a single flight or includes flights that violate the drone range constraint, the node types are adjusted to be serviced by trucks instead.
The offspring is subsequently incorporated into the feasible or infeasible subpopulation based on its feasibility status.
Should the offspring exhibit feasibility, local search \bluenote{neighborhoods} are applied to enhance solution quality (Lines \ref{GAline:local1} and \ref{GAline:local2}), with detailed explanations of local search \bluenote{neighborhoods} provided in Section \ref{improve}.

\begin{algorithm}
    \caption{Hybrid Genetic Algorithm with Type-Aware Chromosomes} \label{alg:GA}
    \begin{algorithmic}[1]
        \State $\Omega_\textsc{f} \gets \emptyset$ \label{GAline:init1} \Comment{Feasible subpopulation}
        \State $\Omega_\textsc{inf}^1 \gets \emptyset$ \Comment{Infeasible subpopulation (Type 1)}  
        \State $\Omega_\textsc{inf}^2 \gets \emptyset$ \Comment{Infeasible \bluenote{subpopulation} (Type 2)}     
        \State $\Omega_\textsc{f}, \Omega_\textsc{inf}^1 , \Omega_\textsc{inf}^2 = \texttt{initial\_population()}$ \label{GAline:init2} \Comment{Algorithm \ref{alg:init}}
        \While{number of iterations with no improvement $<It_{\textsc{NI}}$}  \label{GAline:mainloop1}
            \State Select $\omega_1$ and $\omega_2$ from $\Omega_\textsc{f} \Omega_\textsc{inf}^1 \cup \Omega_\textsc{inf}^2$  \label{GAline:parentselect}
            \State $\omega \gets \texttt{crossover}(\omega_1, \omega_2)$     \label{GAline:crossover}
            \State $\texttt{mutate}(\omega)$ 		\label{GAline:mutation}	%
            \If{$\omega$ is feasible}   \label{GAline:feasible}
                \State $\texttt{local\_search}(\omega)$	  \label{GAline:local1}  %
                \State $\Omega_\textsc{f} \gets \Omega_\textsc{f} \cup \{\omega\}$
            \Else
                \State $r \gets \texttt{rand}(0,1)$    
                \If{$r<P_{\textsc{REPAIR}}$}    \label{GAline:repair1}
                    \State Make $\omega$ feasible   \label{GAline:repair2}
                    \State $\texttt{local\_search}(\omega)$  \label{GAline:local2} 
                    \State $\Omega_\textsc{f} \gets \Omega_\textsc{f} \cup \{\omega\}$
                \Else 
                        \If{$\omega$'s infeasibility is Type 1}
                            \State $\Omega_\textsc{inf}^1 \gets \Omega_\textsc{inf}^1 \cup \{\omega\}$
                        \Else
                            \State $\Omega_\textsc{inf}^2 \gets \Omega_\textsc{inf}^2 \cup \{\omega\}$
                        \EndIf
                \EndIf
            \EndIf 
                \For{$\Omega \in \{\Omega_{\textsc{f}}, \Omega_{\textsc{inf}}^1 , \Omega_{\textsc{inf}}^2 \}$}  \label{GAline:survive1} 
                    \If{ $\texttt{size}(\Omega) = \mu+\lambda $}
                        \State $\texttt{select\_survivors}(\Omega)$
                    \EndIf
                \EndFor  \label{GAline:survive2} 

            \State Adjust penalties   \label{GAline:adjust}
            \If{$\texttt{best}(\Omega_\textsc{f})$ not improved for $It_{\textsc{DIV}}$ iterations}  \label{GAline:diversify1}
                \State $\texttt{diversify}(\Omega_\textsc{f}, \Omega_\textsc{inf}^1 , \Omega_\textsc{inf}^2)$
            \EndIf \label{GAline:diversify2}
            \If{$\texttt{best}(\Omega_\textsc{f})$ not improved for $It_{\textsc{LOC}}$ iterations}  \label{GAline:escape1}
                \State $\texttt{escape\_local\_optima}(\texttt{best}(\Omega_\textsc{f}))$. \Comment{Algorithm \ref{alg:escape}}
            \EndIf \label{GAline:escape2}
    
        \EndWhile \label{GAline:mainloop2}
        \State Return $\texttt{best}(\Omega_\textsc{f})$
    \end{algorithmic}
\end{algorithm}

If any of the subpopulations attain a size of $\mu + \lambda$, the algorithm selectively retains the best $\mu$ solutions while discarding the remainder (Lines \ref{GAline:survive1}-\ref{GAline:survive2}).
Concurrently, the algorithm dynamically manages the generation of infeasible solutions by iteratively adjusting infeasibility penalties (Line \ref{GAline:adjust}) (Section \ref{pop} provides comprehensive details).
In instances where the solution fails to exhibit improvement after $It_{\textsc{DIV}}$ iterations, a population diversification strategy is employed (Lines \ref{GAline:diversify1}-\ref{GAline:diversify2}).
In this step, the first $n_{\textsc{best}}$ individuals are retained and the rest will be removed.
Then new individuals will be generated according to Algorithm \ref{alg:init} until each subpopulation size reaches $\mu$ (Section \ref{pop}).
Additionally, if no enhancements are observed \bluenote{after} $It_{\textsc{LOC}}$ iterations, the algorithm endeavors to escape from local optima using an alternative technique expounded upon in Section \ref{sub:escape} (Lines \ref{GAline:escape1}-\ref{GAline:escape2}). 
It is pertinent to emphasize that $It_{\textsc{DIV}} < It_{\textsc{LOC}} < It_{\textsc{NI}}$.

\subsection{Type-Aware Chromosome Encoding} \label{rep}

The route taken by each vehicle, as well as the locations where the drone is launched from and lands on the truck, are the solutions to the TSPD and the FSTSP. 
Our HGA for TSPD/FSTSP is unique in how we encode each solution in a chromosome, which keeps a sequence of customer nodes and records the type of each node: either a truck node or a drone node.
Hence, we call our HGA the HGA with Type-Aware Chromosomes (HGA-TAC).
While each number in HGA-TAC's Chromosome encoding represents a customer node in the sequence, the truck nodes are represented by positive numbers, and the drone nodes are shown by negative numbers. 
Therefore, each chromosome includes the truck route as well as the drone route, and the TSPD/FSTSP tour will be accomplished if the launch and landing points are known. 
To handle such type-aware chromosomes, we devise a novel DP formulation and type-aware crossover operations.
The proposed DP method, which we call \JOIN{}, determines the optimal launch and landing points for the drone.
The details of the \JOIN{} algorithm are discussed in Section \ref{DP}. 

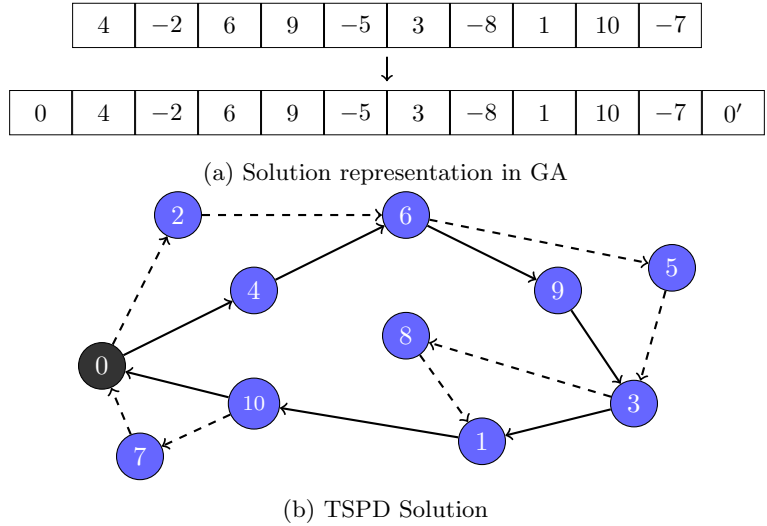
\begin{figure}
\begin{subfigure}[b]{\linewidth}\centering
	\resizebox{4in}{!}{%
		\begin{tikzpicture}
            \matrix[chromosome] (P1) 
            {
               4 \& 
              -2 \& 
               6 \&
               9 \&
              -5 \&
               3 \&
              -8 \&
               1 \&
              10 \&
              -7 \\
            };
            
            \matrix[below = 1em of P1, chromosome] (P2) 
            {
               0 \&
               4 \& 
              -2 \& 
               6 \&
               9 \&
              -5 \&
               3 \&
              -8 \&
               1 \&
              10 \&
              -7 \& 
              0' \\
            };            
            
            	\draw[->, thick] (P1) -- (P2);
		\end{tikzpicture}%
	}
    \caption{Solution representation in GA}
    \label{fig:representation}	
\end{subfigure}
\begin{subfigure}[b]{\linewidth}\centering
	\begin{tikzpicture}
		\node[depotnode] (n0) at (0,0) {0};
		\node[customernode] (n1) at (5,-1) {1};
		\node[customernode] (n2) at (1,2) {2};
		\node[customernode] (n3) at (7,-0.5) {3};
		\node[customernode] (n4) at (2,1) {4};		
		\node[customernode] (n5) at (7.5,1.3) {5};		
		\node[customernode] (n6) at (4,2) {6};				
		\node[customernode] (n7) at (0.5,-1.2) {7};				
		\node[customernode] (n8) at (4,0.4) {8};				
		\node[customernode] (n9) at (6,1) {9};				
		\node[customernode] (n10) at (2,-0.5) {\footnotesize 10};				
				
		\draw[->, thick] (n0) -- (n4);
		\draw[->, thick] (n4) -- (n6);
		\draw[->, thick] (n6) -- (n9);
		\draw[->, thick] (n9) -- (n3);		
		\draw[->, thick] (n3) -- (n1);
		\draw[->, thick] (n1) -- (n10);
		\draw[->, thick] (n10) -- (n0);
		
		\draw[->, thick, dashed] (n0) -- (n2);
		\draw[->, thick, dashed] (n2) -- (n6);
		\draw[->, thick, dashed] (n6) -- (n5);
		\draw[->, thick, dashed] (n5) -- (n3);
		\draw[->, thick, dashed] (n3) -- (n8);
		\draw[->, thick, dashed] (n8) -- (n1);
		\draw[->, thick, dashed] (n10) -- (n7);
		\draw[->, thick, dashed] (n7) -- (n0);
										
	\end{tikzpicture}
	
    \caption{TSPD Solution}
    \label{fig:solution}
\end{subfigure}
\caption{For a network of one depot and 10 customers, the figure in (a) is the solution representation, and (b) is the same solution shown on network.}
\end{figure}

A simple example will serve as a better understanding of our chromosome encoding.
Consider a network that has ten customers.
Figure \ref{fig:representation} illustrates how the chromosome of an individual is coded in our algorithm before and after the addition of the depot. 
As mentioned previously, positive and negative numbers represent the nodes visited by truck and drone, respectively.
The truck tour in this example is $[0,4,6,9,3,1,10,0']$, while the drone tour is $[0,2,5,8,7,0']$. 
The TSPD solution is shown in Figure \ref{fig:solution} if we assume the launch and landing locations found by  \JOIN{} are $\{0,6,3,10\}$ and $\{6,3,1,0'\}$ respectively. 

\subsection{Generating Initial Subpopulations} \label{init-pop}
In the case, \bluenote{like ours}, that the drone is limited to only one customer per flight, any two or more consecutive negative numbers would be considered infeasible \bluenote{(Type 1)} in our chromosome encoding.
Additionally, if the drone's flight range is restricted, a representation can be evaluated for feasibility \bluenote{(Type 2)}. 
\bluenote{Therefore, when the drone range is unlimited, our GA will consist of two subpopulations; otherwise, it will consist of three subpopulations.}

Algorithm \ref{alg:init} illustrates how the initial population is generated. 
Initially, we find a TSP solution, then promote it to a TSPD solution using the exact partitioning proposed in \cite{agatz2018optimization}.
The initial TSP solution is found using the LKH library \citep{helsgaun2000effective}, an efficient and improved implementation of the Lin-Kernighan algorithm \citep{lin1973effective}.
To diversify the population and explore the solution space effectively, new individuals are generated by applying specific modifying operators to existing solutions. 
The process involves the utilization of two distinct operators, each contributing to the exploration of different aspects of the search space.

        \begin{description}	
        \item[Element-Wise Modification:]
        This operator involves the random modification of individual elements within a chromosome. 
        Specifically, for each element in the chromosome, one of the following actions is randomly chosen: sign change, element swap with the successor, or no action. 
        The selection of actions is governed by random numbers in the range $[0, 1]$.
        If the randomly generated number falls within the range $[0, 0.1]$, a sign change is applied to the element.
        In the range $(0.1, 0.2]$, a swap with the succeeding element takes place.
        For numbers beyond $0.2$, no action is executed, leaving the element unaltered (Lines 8-11).

        \item[Sequence Modification:]
        This operator introduces variability by modifying subsequences within the chromosome. 
        Two indices, $i$ and $j$, are randomly selected, and one of the following operations is performed on the sequence between $i$ and $j$ (inclusive): 
        reverse sequence, sign change within the sequence, or shuffle sequence. 
        The choice of operation is determined randomly (Lines 13,14).
        \end{description}

It is imperative to iterate through the aforementioned process until each subpopulation attains a population size of $\mu$.

\begin{algorithm}
    \caption{Initial Population} \label{alg:init}
    \begin{algorithmic}[1]
        \State TSP\_tour $\gets$ LKH(Network)
        \State $\omega_0 \gets$ \EP{}(TSP\_tour)  \Comment{Exact Partitioning by \citet{agatz2018optimization}}
        \State $\Omega_\textsc{f} \gets \{\omega_0\}, \Omega_\textsc{inf}^1 \gets \emptyset, \Omega_\textsc{inf}^2 \gets \emptyset$   \Comment{Feasible and infeasible subpopulations}
        \While{Size of any subpopulation $<\mu$}
            \State $\omega \gets \omega_0$ 
            \State $r \gets \texttt{rand}(0,1) $ 
            \If{$r<0.5$}
                \ForEach{$i \in 1:n$}
                    \State With probability 0.1: $\omega[i] \gets - \omega[i]$ 
                    \State With probability 0.1: $\texttt{swap}(\omega[i], \omega[i+1])$ or $\texttt{swap}(\omega[i], \omega[i-1])$             
				\EndFor
            \Else
                \State Randomly choose $i_1$ and $i_2$ from $1,2,...,n$
                    \State $\omega[i_1:i_2] \gets \omega[i_2:i_1] \text{ or } -\omega[i_1:i_2] \text{ or } \texttt{shuffle}(\omega[i_1:i_2])$
            \EndIf 
            \If{$\omega$ is feasible}
                \State $\Omega_\textsc{f} \gets \Omega_\textsc{f} \cup \{\omega\}$
            \Else
                    \If{$\omega$ is Type 1 infeasible}
                        \State $\Omega_\textsc{inf}^1 \gets \Omega_\textsc{inf}^1 \cup \{\omega\}$
                    \Else 
                        \State $\Omega_\textsc{inf}^2 \gets \Omega_\textsc{inf}^2 \cup \{\omega\}$
                    \EndIf
            \EndIf
        \EndWhile
    \end{algorithmic}
\end{algorithm}

\subsection{Decoding and Evaluating Chromosomes by Dynamic Programming} \label{DP}

The objective of this section is to propose a DP approach, \JOIN{}, to determine the best launch and landing points according to the sequence and type of vehicles used at each node. 
Initially, we will formulate the \JOIN{} algorithm to address sequences that adhere to feasibility constraints. 
Subsequently, we will elucidate the necessary modifications required to extend the algorithm's capabilities in handling various types of infeasible sequences.

As a sub-problem, let us refer to $C(i)$ as the shortest time from truck node $i$ to the return depot. 
By formulating an efficient DP formulation that reduces the entire problem to these subproblems, the optimal objective can be achieved. 
The minimum makespan will be represented by $C(0)$. 
Following is a list of the notation, state transitions, and optimal decisions for a given chromosome being evaluated by \JOIN{}: 

\begin{itemize}
    \item $\tau^\tr_{i \rightarrow k}$: Truck travel time from node $i$ up to node $k$ (visiting all the nodes in between).
    \item $\tau^\dr_{i,j}$: Drone travel time from node $i$ to node $j$. 
    \item $i$: The current truck node in the sub-problem.
    \item $C(i)$: Shortest time of the system of truck and drone from node $i$ to the depot $0'$.
    \item $d(i)$: The closest drone node superseding node $i$ in the chromosome representation (If none, then it would be a dummy node after the depot.)
    \item $d^+(i)$: The drone node superseding $d(i)$ in the chromosome representation (If none, then it would be a dummy node.)
    \item $\set{T}(i)$: The set of truck nodes between $i$ and $d(i)$.
    \item $\set{T}^+(i)$: The set of truck nodes between $d(i)$ and $d^+(i)$.
\end{itemize}

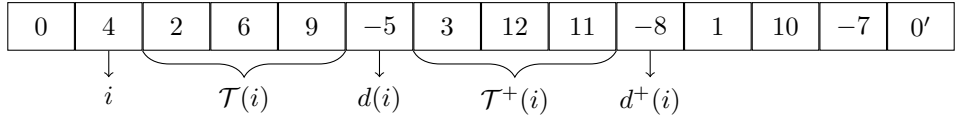
\begin{figure}
	\centering
    \resizebox{5in}{!}{%
        \begin{tikzpicture}
            \matrix [chromosome]
            (A) at (0, 1.3)
            {
              0 \& 4 \& 2 \& 6 \& 9 \& -5 \& 3 \&12 \&11 \& -8 \& 1 \& 10 \& -7 \& 0'  \\
            };
            
            \draw [decorate, decoration={brace, amplitude=10pt, mirror}]
            (A-1-3.south west) -- (A-1-5.south east) node[below=1em, midway] {$\set{T}(i)$};
            
            \draw [decorate, decoration={brace, amplitude=10pt, mirror}]
            (A-1-7.south west) -- (A-1-9.south east) node[below=1em, midway] {$\set{T}^+(i)$};
            
            \draw[->](A-1-2.south) -- +(0, -1em) node[below] {$i$} ;
            \draw[->](A-1-6.south) -- +(0, -1em) node[below] {$d(i)$};
            \draw[->](A-1-10.south) -- +(0, -1em) node[below] {$d^+(i)$};
        
        \end{tikzpicture}%
    }
    \caption{An instance illustrating the notation used in the DP approach}
    \label{fig:DP}    
\end{figure}

For the reader's convenience, Figure \ref{fig:DP} illustrates an example of the representation along with DP notation. 

In each state $i$, the goal is to determine the best move based on the previously computed values and select one of two options, either moving the truck from $i$ to a node in $\set{T}(i)$ while the drone is onboard, or launching the drone from $i$ to visit $d(i)$ and land on a node in $\set{T}^+(i)$ while the truck is visiting all the nodes in between.
For TSPD, the last node of $\set{T}(i)$ must be added to $\set{T}^+(i)$, since the drone is permitted to launch, visit a node, and land, while the truck remains stationary. 
Let us refer to the first type of decision as $\text{MT}$ (move the truck) and the second type as $\text{LL}$ (launch and land).

The transition of states happens as follows: 
\begin{enumerate}
    \item Let $C_{\text{MT}}(i)$ represent the shortest time from node $i$ where the initial decision is to move the truck from $i$ while the drone is onboard. 
    Thus, the required recursion is: 
    \begin{equation*}
        C_\text{MT}(i) = 
        \begin{cases}
             \infty  & \textrm{if } \set{T}(i) = \emptyset, \\
             \displaystyle\min_{k \in \set{T}(i)}
             \big\{ \tau^\tr_{i \rightarrow k} + C(k) \big\} & \textrm{otherwise}.
        \end{cases} 
    \end{equation*}
    \item In the same manner, let $C_{\textsc{LL}}(i)$ be the shortest possible
     time from node $i$, where the first decision is to send the drone
      from $i$ to serve $d(i)$ and then to land on the truck at some node
       in $\set{T}^+(i)$. 
       Based on the different assumptions in TSPD and FSTSP, we must have different functions to calculate $C_{\textsc{LL}}(i)$. 
       For TSPD, let $\mathcal{E}^+(i) = \{k \in \set{T}^+(i): \tau^\dr_{i,d(i)}+\tau^\dr_{d(i),k}\leq e\}$, which is a subset of $\set{T}^+(i)$ and every node in it can form a feasible drone operation with $i$ and $d(i)$. Therefore, the required recursion for TSPD will be:
    \begin{equation*}
        C_{\textsc{LL}}(i) =  
        \begin{cases}
             \infty  & \textrm{if } \mathcal{E}^+(i) = \emptyset, \\
             \displaystyle\min_{k \in \mathcal{E}^+(i)}
             \Big\{
             	\max\big\{
					\tau^\tr_{i \rightarrow k}, \ 
	             	\tau^\dr_{i, d(i)} + \tau^\dr_{d(i), k}
				\big\} + C(k) 
             \Big\} & \textrm{otherwise}.
        \end{cases} 
    \end{equation*}

    For FSTSP, on the other hand, we need to know whether a drone launch has occurred in node $k$. 
    Therefore, let $\sigma_k$ be:
    \begin{equation*}
        \sigma_k =  
        \begin{cases}
             1 & \textrm{if the drone is launched at node $k$}\\
             0 & \textrm{otherwise}.
        \end{cases} 
    \end{equation*}
    Now we define $\mathcal{E}^+(i) = \{k \in \set{T}^+(i): \tau^\tr_{i \rightarrow k} +s_R + \sigma_k s_L \leq e, \ \tau^\dr_{i,d(i)}+\tau^\dr_{d(i),k} + s_R \leq e\}$. 
    Therefore, the required recursion for FSTSP will be:
    
        \begin{equation*}
        C_{\textsc{LL}}(i) =  
        \begin{cases}
             \infty  & \textrm{if } \mathcal{E}^+(i) = \emptyset, \\
             \displaystyle\min_{k \in \mathcal{E}^+(i)} 
             \Big\{
             	\max\big\{
					\tau^\tr_{i \rightarrow k} + s_R + \sigma_k s_L, \ 
		            \tau^\dr_{i, d(i)} + \tau^\dr_{d(i), k} + s_R
		        \big\} + C(k) 
		     \Big\} & \textrm{otherwise}.
        \end{cases} 
    \end{equation*}

    \item Finally the required recursion for state transition is: 
    \begin{equation*}
        C(i) = \min\big\{C_{\textsc{MT}}(i), \  C_{\textsc{LL}}(i)\big\},
    \end{equation*}
    where $C(0')=0$. Therefore, the \JOIN{} algorithm finds the rendezvous points as well as the minimum total time through backward recursion.  
    Note that $\mathcal{E}^+(i) = \emptyset$ and $\set{T}(i) = \emptyset$ would never happen simultaneously.
    Therefore, $C(i)$ will always have a finite value for any state $i$. 
\end{enumerate}

    Following the detailed explanation of the algorithm, we present a formal summary of the \JOIN{} algorithm tailored for feasible TSPD sequences in Algorithm \ref{alg:join}. 
    The algorithm ensures optimality through a process of backward propagation, wherein, at each stage, it systematically identifies the minimum makespan from node $i$ to the return depot node by examining all feasible movements.
    Additionally, it is noteworthy that Algorithm \ref{alg:join} can be readily adapted to accommodate feasible FSTSP sequences by substituting the equation for $C_{\textsc{LL}}(i)$. 
    Subsequent instructions on the necessary modifications to enable \JOIN{} to handle infeasible sequences will be provided shortly.

\begin{algorithm}
    \caption{\JOIN{} algorithm for finding optimal launch and land locations.} \label{alg:join}
    \begin{algorithmic}[1]
        \State Set of all truck nodes: $\{0, i_1, i_2, ..., i_m, 0'\}$ ($m \leq n$)
        \State $C(i) = \infty , C_\text{MT}(i) = \infty, C_\text{LL}(i) = \infty$ for $i = 0, i_1, i_2, ..., i_m$ \Comment{Initialization}
        \State $C(0') = 0$ 
        \For{$i=i_m, i_{m-1}, ... , i_1, 0$}
            \For{$k \in \set{T}(i)$}
                \If{$\tau^\tr_{i \rightarrow k} + C(k) < C_\text{MT}(i)$}
                    \State$C_\text{MT}(i) = \tau^\tr_{i \rightarrow k} + C(k)$
                \EndIf
            \EndFor
            \For{$k \in \mathcal{E}^+(i)$}
                \If{$\max\big\{ \tau^\tr_{i \rightarrow k}, \tau^\dr_{i, d(i)} + \tau^\dr_{d(i), k}	\big\} + C(k)  < C_\text{LL}(i)$}
                    \State$C_\text{LL}(i) = \max\big\{ \tau^\tr_{i \rightarrow k}, \tau^\dr_{i, d(i)} + \tau^\dr_{d(i), k}	\big\} + C(k)$
                \EndIf
            \EndFor
            \State $C(i) = \min \{C_\text{MT}(i), C_\text{LL}(i)\}$
        \EndFor
        \State return $C(0)$
    \end{algorithmic}
\end{algorithm}

    The time complexity analysis of the algorithm reveals an efficient computational structure. 
    The initial loop traverses all truck nodes, denoted by $m$, where $m \leq n$. 
    Within this primary loop, two nested loops are implemented. 
    The first inner loop iterates over $k \in \set{T}(i)$, where $|\set{T}(i)| < m$. 
    Simultaneously, the second inner loop iterates over $k \in \mathcal{E}^+(i)$, where $|\mathcal{E}^+(i)| \leq |\set{T}^+(i)| < m$.
    As a result of this structured arrangement, the worst-case time complexity of the algorithm is $O(n^2)$. 
    This complexity arises from the combination of traversing all truck nodes, the limited cardinality of the sets $\set{T}(i)$ and $\mathcal{E}^+(i)$, ensuring an efficient and scalable computational performance even in larger problem instances.

\begin{lemma}
By using the \JOIN{} algorithm for a given chromosome where the sequence and types of the vehicles are known, TSPD (or FSTSP) solutions can be determined in time $O(n^2)$. 
\end{lemma}

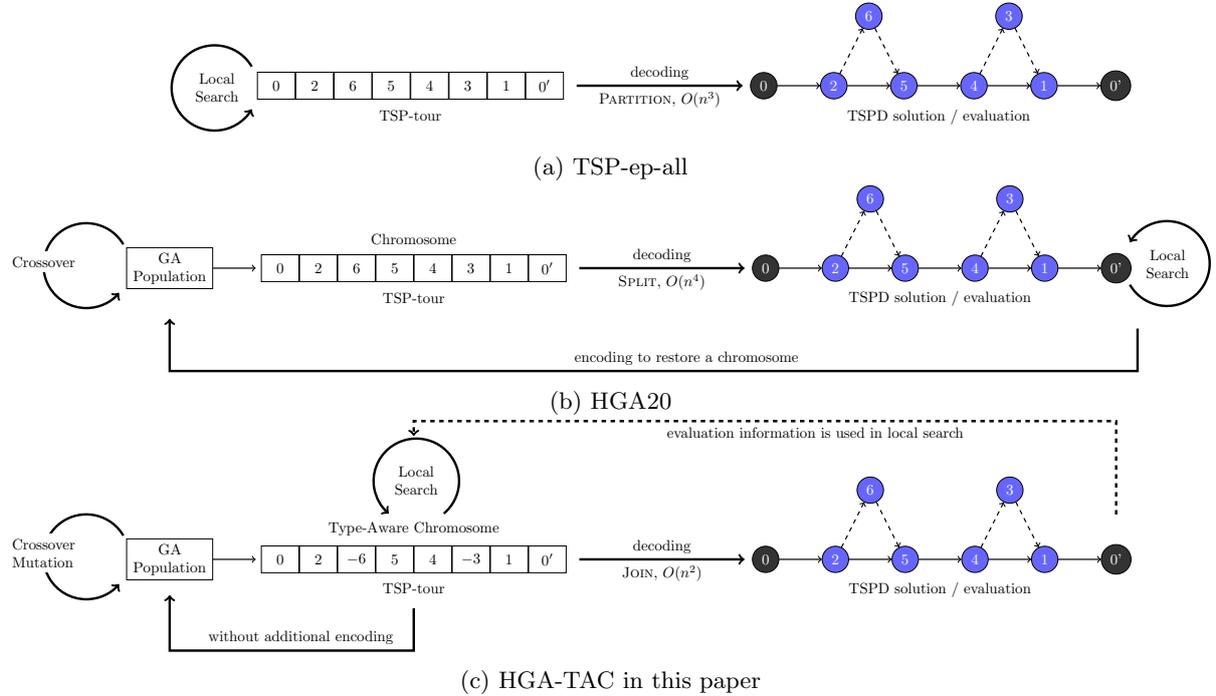
\begin{figure}
    \centering
    \begin{subfigure}[b]{\linewidth}\hfill
        \resizebox{0.87\textwidth}{!}{%
            \begin{tikzpicture}
                    
            \matrix[chromosome, label=below:TSP-tour] (A1)
            {
                0 \& 
                2 \& 
                6 \& 
                5 \& 
                4 \& 
                3 \& 
                1 \& 
                0'\\ 
            };
            \node[depotnode, right=120pt of A1] (n0) {0};
            \node[customernode, right=of n0] (n2) {2};
            \node[customernode, right=of n2] (n5) {5};
            \node[customernode, right=of n5] (n4) {4};
            \node[customernode, right=of n4] (n1) {1};
            \node[depotnode, right=of n1] (n7) {0'};
            \node[customernode, above=of $(n2.north)!0.5!(n5.north)$] (n6) {6};
            \node[customernode, above=of $(n4.north)!0.5!(n1.north)$] (n3) {3};

            \draw[->, thick] (n0) -- (n2);
            \draw[->, thick] (n2) -- (n5);
            \draw[->, thick] (n5) -- (n4);
            \draw[->, thick] (n4) -- (n1);
            \draw[->, thick] (n1) -- (n7);
            \draw[->, thick, dashed] (n2) -- (n6);
            \draw[->, thick, dashed] (n6) -- (n5);
            \draw[->, thick, dashed] (n4) -- (n3);
            \draw[->, thick, dashed] (n3) -- (n1);
            \draw[->, ultra thick] ([xshift=5pt]A1.east) -- node[above] {decoding} node[below] {\EP{}, $O(n^3)$} ([xshift=-5pt]n0.west);
            \draw[ultra thick, ->] (A1.north west) arc (30:330:1) node[swap, fill=white, midway, align=center, xshift=30pt] {Local\\Search};
            \node[below=3pt of $(n5.south)!0.5!(n4.south)$] {TSPD solution / evaluation};

            \draw[ultra thick, ->, draw=white] ([xshift=2pt, yshift=-4pt]n7.south east) arc (210:510:1) node[swap, fill=white, midway, align=center, xshift=-28pt, text=white] (ls) {Local\\Search};

            \end{tikzpicture}%
        }            
    \caption{TSP-ep-all}        
    \end{subfigure}

    \begin{subfigure}[b]{\linewidth}\hfill
        \resizebox{\textwidth}{!}{%
            \begin{tikzpicture}
                
            \node[draw, text width = 0.7in, align = center] (GA) {GA\\ Population};
            
            \matrix[chromosome, label=below:TSP-tour, label=above:Chromosome, right=of GA] (A1)
            {
                0 \& 
                2 \& 
                6 \& 
                5 \& 
                4 \& 
                3 \& 
                1 \& 
                0'\\ 
            };
            \draw[->, thick] (GA) -- (A1);
            \node[depotnode, right=120pt of A1] (n0) {0};
            \node[customernode, right=of n0] (n2) {2};
            \node[customernode, right=of n2] (n5) {5};
            \node[customernode, right=of n5] (n4) {4};
            \node[customernode, right=of n4] (n1) {1};
            \node[depotnode, right=of n1] (n7) {0'};
            \node[customernode, above=of $(n2.north)!0.5!(n5.north)$] (n6) {6};
            \node[customernode, above=of $(n4.north)!0.5!(n1.north)$] (n3) {3};

            \draw[->, thick] (n0) -- (n2);
            \draw[->, thick] (n2) -- (n5);
            \draw[->, thick] (n5) -- (n4);
            \draw[->, thick] (n4) -- (n1);
            \draw[->, thick] (n1) -- (n7);
            \draw[->, thick, dashed] (n2) -- (n6);
            \draw[->, thick, dashed] (n6) -- (n5);
            \draw[->, thick, dashed] (n4) -- (n3);
            \draw[->, thick, dashed] (n3) -- (n1);
            \draw[->, ultra thick] ([xshift=5pt]A1.east) -- node[above] {decoding} node[below] {\SPLIT{}, $O(n^4)$} ([xshift=-5pt]n0.west);
            \draw[ultra thick, ->] ([xshift=-2pt, yshift=2pt]GA.north west) arc (30:320:1) node[swap, fill=white, midway] {Crossover};
            \draw[ultra thick, ->] ([xshift=2pt, yshift=-4pt]n7.south east) arc (210:510:1) node[swap, fill=white, midway, align=center, xshift=-28pt] (ls) {Local\\Search};
            \node[below=3pt of $(n5.south)!0.5!(n4.south)$] {TSPD solution / evaluation};
            \draw[->, ultra thick] ([xshift=-20pt, yshift=-30pt]ls.south) |- ++(0, -1) 
            -- node[above]{encoding to restore a chromosome} ++(-600pt, 0) -| ([yshift=-20pt]GA.south);

            \end{tikzpicture}%
        }
        \caption{HGA20}
    \end{subfigure}
    \begin{subfigure}[b]{\linewidth}\hfill
        \resizebox{\textwidth}{!}{%
            \begin{tikzpicture}
                    
            \node[draw, text width = 0.7in, align = center] (GA) {GA\\ Population};
            
            \matrix[chromosome, label=below:TSP-tour, label=above:Type-Aware Chromosome, right=of GA] (A1)
            {
                0 \& 
                2 \& 
                -6 \& 
                5 \& 
                4 \& 
                -3 \& 
                1 \& 
                0'\\ 
            };
            \draw[->, thick] (GA) -- (A1);
            \node[depotnode, right=120pt of A1] (n0) {0};
            \node[customernode, right=of n0] (n2) {2};
            \node[customernode, right=of n2] (n5) {5};
            \node[customernode, right=of n5] (n4) {4};
            \node[customernode, right=of n4] (n1) {1};
            \node[depotnode, right=of n1] (n7) {0'};
            \node[customernode, above=of $(n2.north)!0.5!(n5.north)$] (n6) {6};
            \node[customernode, above=of $(n4.north)!0.5!(n1.north)$] (n3) {3};

            \draw[->, thick] (n0) -- (n2);
            \draw[->, thick] (n2) -- (n5);
            \draw[->, thick] (n5) -- (n4);
            \draw[->, thick] (n4) -- (n1);
            \draw[->, thick] (n1) -- (n7);
            \draw[->, thick, dashed] (n2) -- (n6);
            \draw[->, thick, dashed] (n6) -- (n5);
            \draw[->, thick, dashed] (n4) -- (n3);
            \draw[->, thick, dashed] (n3) -- (n1);
            \draw[->, ultra thick] ([xshift=5pt]A1.east) -- node[above] {decoding} node[below] {\JOIN{}, $O(n^2)$} ([xshift=-5pt]n0.west);
            \draw[ultra thick, ->] ([xshift=-2pt, yshift=2pt]GA.north west) arc (30:320:1) node[swap, fill=white, midway, align=left] {Crossover\\Mutation};
                \node[below=3pt of $(n5.south)!0.5!(n4.south)$] {TSPD solution / evaluation};

            \draw[->, ultra thick] ([yshift=-20pt]A1.south) |- ++(0, -1) 
            -- node[above]{without additional encoding} ++(-150pt, 0) -| ([yshift=-10pt]GA.south);

            \draw[->, dashed, ultra thick] ([yshift=20pt]n7.north) |- ++(0, 2.2) 
            -- node[below]{evaluation information is used in local search} ++(-400pt, 0) -| ([yshift=70pt]A1.north);

            \draw[ultra thick, ->] ([xshift=20pt, yshift=18pt]A1.north) arc (-50:230:1) node[swap, fill=white, midway, align=center, yshift=-28pt] (ls) {Local\\Search};
            
            \draw[ultra thick, ->, draw=white] ([xshift=2pt, yshift=-4pt]n7.south east) arc (210:510:1) node[swap, fill=white, midway, align=center, xshift=-28pt, text=white] (ls) {Local\\Search};

            \end{tikzpicture}%
        }
    \caption{HGA-TAC in this paper}
    \label{fig:HGA-TAC}
    \end{subfigure}
    \caption{Comparing the structure of our approach to TSP-ep-all in \citet{agatz2018optimization} and HGA20 in \citet{ha2020hybrid}}
    \label{fig:comparison} 
\end{figure}

With the aim of demonstrating the importance of \JOIN{}'s low time complexity, we now seek to conduct a comparative examination of our algorithm's structure in relation to TSP-ep-all and HGA20. 
To facilitate this comparison, Figure \ref{fig:comparison} presents a comprehensive depiction of our algorithm's architecture, allowing for a thorough analysis and contrasting of its attributes with those of TSP-ep-all and HGA20.
Observably, TSP-ep-all takes a TSP tour and applies exact partitioning, or \EP{}, to transform it optimally to a TSPD solution with time complexity of $O(n^3)$ \citep{agatz2018optimization}, $n$ being the number of customer nodes.
In HGA20, multiple TSP tours are generated by crossover within GA, and the TSPD solution is obtained using the \SPLIT{} method with a time complexity of $O(n^4)$ \citep{ha2018min}, which is analogous to \EP{}.
Then the TSPD solution is restored as a chromosome (TSP tour) and reintroduced to the GA population pool. 

In our methodology, each chromosome is designed to store more information. 
GA also determines the vehicle type used to serve each customer, in addition to the order. 
As illustrated in Figure \ref{fig:HGA-TAC}, each chromosome consists of a sequence of numbers that represent the customers. 
The positive and negative numbers indicate the customers that will be visited by truck and drone, respectively.  
To achieve the TSPD solution, it is sufficient to establish where the drone is released from and returns to the truck.
The \JOIN{} algorithm is employed to discover the \bluenote{optimal} rendezvous points, with the time complexity of $O(n^2)$. 
To summarize, the type of vehicle required for each node is decided in \EP{} and \SPLIT{}, in addition to rendezvous points, while in our \JOIN{} approach, the first decision has been assigned to GA. 

There is another point of comparison that can be drawn between these three approaches: local search.
The TSP-ep-all algorithm executes local search on TSP tours before applying \EP{}, while the HGA20 algorithm performs local search after applying \SPLIT{} on TSPD solutions before converting back to chromosomes. 
Our approach integrates these techniques directly onto the chromosome.
Unlike HGA20, which necessitates a conversion back to the chromosome after local search, our method streamlines the process by conducting the local search directly on the chromosome. 
This modification eliminates the need for an additional encoding step.

\subsection*{Evaluation of infeasible sequences}

It is essential to emphasize that the \JOIN{} \bluenote{procedure} presented in Algorithm \ref{alg:join} is intended to compute the shortest possible time for feasible solutions. 
Given that infeasible solutions are ultimately unfavorable, they must be penalized during the evaluation process. 
The \JOIN{} algorithm will be able to calculate the makespan for infeasible solutions with some slight modifications, denoted as penalized makespan.
As previously stated, we exploit two types of infeasibility. 
For representations with at least two adjacent negative nodes, the drone violates the premise of a single visit per fly. 
Let $w_1 > 1$ be the penalty for this type of infeasibility (Type 1). 
The travel time for drone launching at $i$, visiting $j_1,j_2,...,j_l$ and landing at $k$, will be calculated as
    $\tau^\dr_{i,j_1}+\sum_{q=1}^{l-1}w_1^q \tau^\dr_{j_{q},j_{q+1}}+\tau^\dr_{j_l,k}$. 
    
On the other hand, if the flying range is violated by the drone in TSPD or by any of the vehicles in FSTSP, the solution is Type 2 infeasible.
With only a few modifications, the same recursions can be used to calculate the cost.
To begin with, set $\mathcal{E}^+(i) $ should be replaced with set $\set{T}^+(i)$, since all movements are possible, regardless of whether or not the drone range constraint is violated. 
Let $w_2$ be the penalty for Type 2 infeasibility.
For TSPD we only need to add $w_2 \max\{0,\tau^\dr_{i,d(i)}+\tau^\dr_{d(i),k} - e\}$ to $\tau^\dr_{i,d(i)}+\tau^\dr_{d(i),k}$.
For FSTSP we need to add $w_2 \max\{0,\tau^\tr_{i \rightarrow k} + s_R + \sigma_k s_L - e\}$ to $\tau^\tr_{i \rightarrow k} + s_R + \sigma_k s_L $ and add $w_2 \max\{0,\tau^\dr_{i,d(i)} +\tau^\dr_{d(i),k} + s_R  - e\}$ to $\tau^\dr_{i,d(i)} +\tau^\dr_{d(i),k} + s_R$.

With the introduction of infeasibility considerations into Algorithm \ref{alg:join}, it is noteworthy that the modifications exclusively impact the equation associated with $C_\text{LL}(i)$, leaving the overall structure of the algorithm unaltered. 
Consequently, the time complexity remains unchanged at $O(n^2)$, emphasizing the continued efficiency and scalability of the algorithm even in the presence of infeasible sequences.

\subsection*{Fitness evaluation}

The optimal objective value, as determined by the \JOIN{} algorithm, is the makespan obtained by the TSPD/FSTSP solution for a single feasible individual (penalized makespan if infeasible).
In order to have greater diversity, the fitness measure will be calculated for each individual as a combination of the makespan (penalized makespan) and \bluenote{a} diversity factor. 
Following \citet{vidal2012hybrid}, for two individuals $P_1$ and $P_2$, a normalized Hamming distance $\delta^H(P_1, P_2)$ is defined as following: 
\[\delta^H(P_1, P_2)=\frac{1}{n}\sum_{i=1}^{n}\boldsymbol{1}(P_1[i]\ne P_2[i]),\]
where $\boldsymbol{1}(\cdot)$ is equal to 1 if the condition specified within the parentheses is true and 0 otherwise, and $n$ is the number of customers. 
The distance ranges between zero and one, with one indicating two individuals are completely different and zero indicating they are the same representation. 
The population is sorted according to the makespan determined by the \JOIN{} algorithm, and diversity contribution $\Delta(P)$ is calculated by the average Hamming distance between an individual $P$ and its two closest individuals within the population. 
According to the following formula, the fitness function for each individual is calculated:
\[\text{fitness}(P) = C_{\max}(P) \times \Big(1-\frac{n_{\textsc{Elite}}}{n_{\textsc{Population}}} \Big)^{\Delta(P)},\]
where we let $C_{\max}(P)$ be the optimal makespan by the \JOIN{} algorithm given individual $P$ (and penalized makespan if $P$ is infeasible), $n_{\textsc{Elite}}$ is the number of elite individuals and $n_{\textsc{Population}}$ is the total number of individuals in the population.

This equation presents a nuanced modification compared to the formulation introduced by \citet{vidal2012hybrid}. 
In our adaptation, the diversity factor serves as the exponent, whereas in \citet{vidal2012hybrid}'s work, it is employed as a multiplier.

\subsection{Parent Selection, Crossover, and Mutation} \label{parent}
During the crossover process, we select two parents, $P_1$ and $P_2$, from which a new individual is produced.
Various methods are available in the literature for selecting parents.
For this study, we employ the \emph{Tournament Selection} in which $k_{\textsc{TOURNAMENT}}$ individuals are randomly selected from the entire population, and the best one is selected as the parent based on fitness. 
A repeat of this process is conducted until two parents have been selected.
It is important to note that before selecting parents, the entire population must be sorted according to fitness, introduced in Section \ref{DP}, to accommodate diversity in the generation of offspring.

Several crossover methods designed for TSP for the generation of offspring exist in the literature. 
Having performed numerical experiments, we randomly employ four crossover methods, including {Order Crossover} (OX1) and {Order-based Crossover} (OX2) with minor modifications adapted to fit our problem and two crossovers designed specifically for TSPD in this study. 
\citet{larranaga1999genetic} contains details on these crossover methods, which is a review paper describing different representations and operators of GA for TSP.
By creating two random crossover points in the parent, OX1 copies the segment between these crossover points to the offspring.
From the second crossover point, the remaining unused numbers are copied from the second parent to the offspring in the same order in which they appear in the first parent. 
Upon reaching the end of the parent string, we begin at its first position.
Using the OX2 operator, several positions in a parent string are randomly selected, and the ordered elements in the selected positions in this parent are imposed on the other parent. 
The elements missing from the offspring are added in the same order as in the second parent.

\begin{figure}
	\centering
\begin{subfigure}[b]{\linewidth}	
    \resizebox{\textwidth}{!}{%
        \begin{tikzpicture}
            \matrix[chromosome] (P1) at (0, 1.3)
            {
              -1 \& 
              |[fill=white!80!black]| 2 \& 
              |[fill=white!80!black]| 3 \& 
              -4 \& 
              |[fill=white!80!black]| 5 \& 
			  |[fill=white!80!black]| 6 \& 
              -7 \& 
			   8 \& 
              -9 \& 
              10 \\
            };
            
			\node [above = 0.3em of P1-1-2.north west] {$i_1$};
			\node [above = 0.3em of P1-1-7.north east] {$i_2$};
            \draw [decorate, decoration={brace, amplitude=10pt}]
            (P1-1-2.north west) -- (P1-1-7.north east) node[above=1em, midway] {truck nodes};

            \matrix[right = of P1, chromosome] (P2)
            {
              |[fill=white!80!blue ]| 4 \& 
              -2 \& 
               6 \& 
              |[fill=white!80!blue ]| 9 \& 
              -5 \& 
			   3 \& 
              |[fill=white!80!blue ]|-8 \& 
              |[fill=white!80!blue ]| 1 \& 
              |[fill=white!80!blue ]|10 \& 
              |[fill=white!80!blue ]|-7 \\
            };

            \matrix[below = 4em of $(P1)!0.5!(P2)$, chromosome] (C)
            {
              |[fill=white!80!blue ]| 4 \& 
              |[fill=white!80!black]| 2 \& 
              |[fill=white!80!black]| 3 \& 
              |[fill=white!80!blue ]| 9 \& 
              |[fill=white!80!black]| 5 \& 
			  |[fill=white!80!black]| 6 \& 
              |[fill=white!80!blue ]|-8 \& 
              |[fill=white!80!blue ]| 1 \& 
              |[fill=white!80!blue ]|10 \& 
              |[fill=white!80!blue ]|-7 \\
            };  
            
            	\draw[->] (P1-1-2.south west) -- (C-1-2.north west);
            	\draw[->] (P1-1-7.south east) -- (C-1-7.north east);
        \end{tikzpicture}%
        	}
	\caption{TOX1}        	
\end{subfigure}
\begin{subfigure}[b]{\linewidth}
    \resizebox{\textwidth}{!}{%
        \begin{tikzpicture}
            \matrix[chromosome] (P1) at (0, 1.3)
            {
              -1 \& 
               2 \& 
              |[fill=white!80!black]| 3 \& 
              |[fill=white!80!black]|-4 \& 
              |[fill=white!80!black]| 5 \& 
			  |[fill=white!80!black]| 6 \& 
              |[fill=white!80!black]|-7 \& 
			   8 \& 
              -9 \& 
              10 \\
            };
            
			\node [above = 0.3em of P1-1-3.north west] {$i_1$};
			\node [above = 0.3em of P1-1-7.north east] {$i_2$};
            \draw [decorate, decoration={brace, amplitude=10pt}]
            (P1-1-3.north west) -- (P1-1-7.north east) node[above=1em, midway] {};

            \matrix[right = of P1, chromosome] (P2)
            {
              4 \& 
              |[fill=white!80!blue ]|-2 \& 
               6 \& 
              |[fill=white!80!blue ]| 9 \& 
              -5 \& 
			   3 \& 
              |[fill=white!80!blue ]| -8 \& 
              |[fill=white!80!blue ]| 1 \& 
              |[fill=white!80!blue ]|10 \& 
              -7 \\
            };

            \matrix[below = 1mm of P2, chromosome, minimum height=1em] (S2)
            {
              |[fill=white!80!black]| + \& 
              - \& 
              |[fill=white!80!black]| + \& 
              + \& 
              |[fill=white!80!black]| - \& 
			  |[fill=white!80!black]| + \& 
              - \& 
              + \& 
              + \& 
              |[fill=white!80!black]| - \\
            };  
			\matrix[below = 1mm of P1, chromosome, minimum height=1em] (S1)
            {
              |[fill=white!80!blue ]| - \& 
              |[fill=white!80!blue ]| + \& 
              |[fill=white!80!blue ]| + \& 
              - \& 
              + \& 
			  + \& 
              - \& 
              |[fill=white!80!blue ]| + \& 
              - \& 
              |[fill=white!80!blue ]| + \\
            };  
            
            \matrix[below = 6em of $(P1)!0.5!(P2)$, chromosome] (C)
            {
              |[fill=white!80!blue ]|+2 \& 
              |[fill=white!80!blue ]|-9 \& 
              |[fill=white!80!black]|+3 \& 
              |[fill=white!80!black]|+4 \& 
              |[fill=white!80!black]|-5 \& 
			  |[fill=white!80!black]|+6 \& 
              |[fill=white!80!black]|-7 \& 
              |[fill=white!80!blue ]|+8 \& 
              |[fill=white!80!blue ]|-1 \& 
              |[fill=white!80!blue ]|+10 \\
            };  
            
            	\draw[->] (P1-1-3.south west) -- (C-1-3.north west);
            	\draw[->] (P1-1-7.south east) -- (C-1-7.north east);
	
	\end{tikzpicture}%
	}
	\caption{TOX2}
\end{subfigure}
    \caption{Examples of the type-aware order crossover operations, TOX1 and TOX2}
    \label{fig:crossover}
\end{figure}
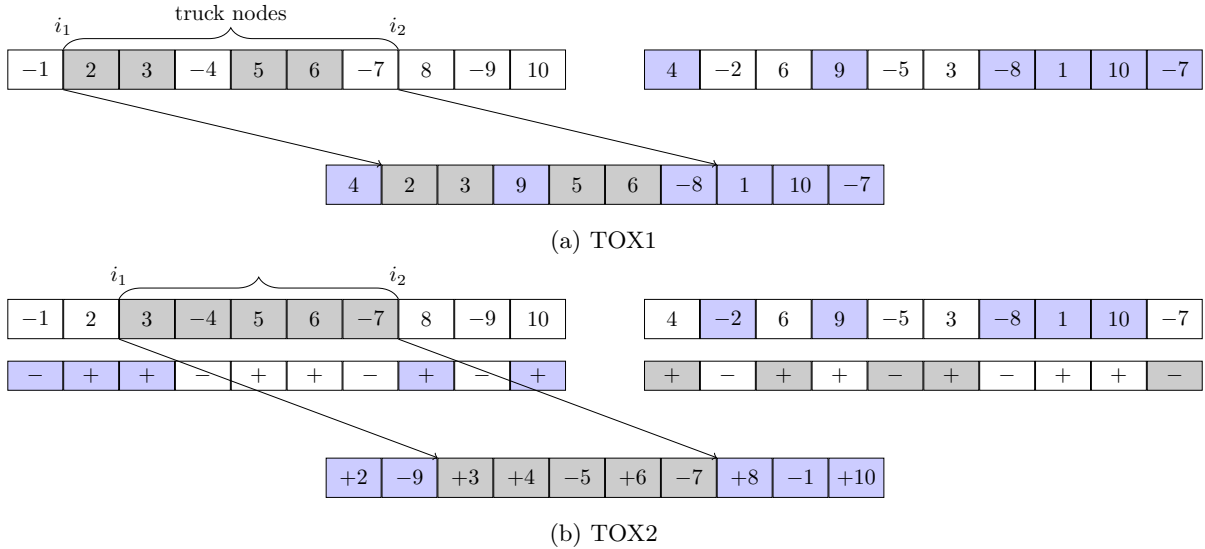

In our modification of OX1 and OX2, we perform these crossovers with absolute values and then retrieve the signs. 
This modification aims to capture both the order and the sign of nodes in the sequence. 
However, despite this enhancement, the resulting crossovers still fall short in fully addressing the unique attributes of TSPD, where the distinction between drone nodes and truck nodes is crucial.

To bridge this gap, we introduce two new crossover algorithms, collectively referred to as Type-Aware Order Crossover (TOX) methods. 
These methods are specifically tailored for TSPD and consider the different types of nodes present in the problem.
In the first algorithm (TOX1), we begin by selecting either truck nodes or drone nodes from $P_1$ and copying the nodes between two randomly chosen indices to the offspring. 
The missing nodes are then added to the offspring in the order and type in which they appear in $P_2$. 
In the second algorithm (TOX2), nodes from $P_1$ are copied into offspring between randomly selected nodes $i_1$ and $i_2$, and the missing nodes are added in the order of $P_2$.
Afterward, the types (signs) of the nodes are determined based on $P_2$ for those between $i_1$ and $i_2$, and based on $P_1$ for the remainder. 
We provide an example of TOX1 and TOX2 processes in Figure \ref{fig:crossover}. 

The motivation for developing TOX1 and TOX2 arises from the need to create crossovers that can capture the differences between drone nodes and truck nodes, ensuring that the offspring inherit all problem-specific properties from both parents. 
By considering the genetic attributes that determine the type of nodes as well as their sequence, TOX1 and TOX2 offer a more nuanced and problem-specific approach to crossover in the context of TSPD.

Empirical studies conducted across various problem domains utilizing GAs have substantiated the effectiveness of applying mutation on offspring.
In our algorithm, once an offspring is produced, it undergoes mutation with a probability of $P_\textsc{m}$.

    Two distinct mutation operators are randomly applied to the offspring: Sign Mutation and Tour Mutation.   
    \begin{description}
	\item[Sign Mutation:] This operator independently changes the sign of each element in the chromosome with a fixed probability of $0.1$.
    \item[Tour Mutation:] In this mutation scheme, $20$ percent of the indices in the chromosome are randomly selected, and the elements at these positions are shuffled. This operation is analogous to the Scramble Mutation \citep{larranaga1999genetic}.
    \end{description}

\subsection{Improvement by Local Search} \label{improve}
In contrast to the traditional GA, hybrid GA algorithms improve each offspring by local search.
\citet{ha2020hybrid} used sixteen local search neighborhoods specific to TSPD/FSTSP, which they call $N_1$ to $N_{16}$.
Our study utilizes 15 of these neighborhoods $N_1$ to $N_{15}$.
Since the \bluenote{neighborhood} $N_{16}$ is randomly modifying launch and rendezvous points for drone deliveries, it will not make any contributions because the \JOIN{} algorithm is optimizing the launch and landing locations.
    
\begin{figure}
	\centering
    \resizebox{\textwidth}{!}{%
        \begin{tikzpicture}
                
            \matrix[chromosome] (A1) at (0, 1.3)
            {
               4 \& 
              -2 \& 
               6 \& 
              |[fill=white!80!black]| 9 \& 
              |[fill=white!80!black]| 5 \& 
              |[fill=white!80!black]| 3 \& 
              -8 \& 
               1 \& 
              10 \& 
              -7 \\
            };
            
            \matrix[right = 1in of A1, chromosome] (B1)
            {
               4 \& 
              -2 \& 
               6 \& 
              |[fill=white!80!black]| 9 \& 
              |[fill=white!80!blue ]| \bf-5 \& 
              |[fill=white!80!black]| 3 \& 
              -8 \& 
               1 \& 
              10 \& 
              -7 \\
            };       

			\draw[->] (A1) -- node[above] {$L_1$} (B1) ;

            \matrix[chromosome, below = 1em of A1] (A2)
            {
               4 \& 
              |[fill=white!80!black]|-2 \& 
               6 \& 
               9 \& 
              |[fill=white!80!blue ]| 5 \& 
              |[fill=white!80!blue ]| 3 \& 
              -8 \& 
               1 \& 
              10 \& 
              -7 \\
            };
            
            \matrix[right = 1in of A2, chromosome] (B2)
            {
               4 \& 
               6 \& 
               9 \& 
              |[fill=white!80!blue ]| 5 \& 
              |[fill=white!80!black]|-2 \& 
              |[fill=white!80!blue ]| 3 \& 
              -8 \& 
               1 \& 
              10 \& 
              -7 \\
            }; 

			\draw[->] (A2) -- node[above] {$L_2$} (B2) ;
		
            \matrix[chromosome, below = 1em of A2] (A3)
            {
               4 \& 
              |[fill=white!80!black]|-2 \& 
               6 \& 
               9 \& 
               5 \& 
              |[fill=white!80!blue]| 3 \& 
              -8 \& 
               1 \& 
              10 \& 
              -7 \\
            };
            
            \matrix[right = 1in of A3, chromosome] (B3)
            {
               4 \& 
              |[fill=white!80!blue]|\bf-3 \& 
               6 \& 
               9 \& 
               5 \& 
              |[fill=white!80!black]|\bf 2 \& 
              -8 \& 
               1 \& 
              10 \& 
              -7 \\
            };

			\draw[->] (A3) -- node[above] {$L_3$} (B3) ;

            \matrix[chromosome, below = 1em of A3] (A4)
            {
               4 \& 
              -2 \& 
              |[fill=white!80!black]| 6 \& 
              |[fill=white!80!black]| 9 \& 
               5 \& 
               3 \& 
              |[fill=white!90!blue]|-8 \& 
              |[fill=white!80!blue]| 1 \& 
              |[fill=white!80!blue]|10 \& 
              -7 \\
            };
            
            \matrix[right = 1in of A4, chromosome] (B4)
            {
               4 \& 
              -2 \& 
              |[fill=white!80!black]| 6 \& 
              |[fill=white!80!blue]| 1 \& 
              |[fill=white!90!blue]|-8 \& 
               3 \& 
               5 \& 
              |[fill=white!80!black]| 9 \& 
              |[fill=white!80!blue]|10 \& 
              -7 \\
            };

			\draw[->] (A4) -- 
				node[above] {$L_4$} 
				node[below] {\footnotesize$\colorbox{white!80!black}{(6,9)} - \colorbox{white!80!blue}{(1,10)}$} 
				(B4) ;
		
            \matrix[chromosome, below = 1em of A4] (A5)
            {
               4 \& 
              |[fill=white!80!black]|-2 \& 
			   6 \& 
               9 \& 
               5 \& 
               3 \& 
              |[fill=white!80!blue]|-8 \& 
               1 \& 
              10 \& 
              -7 \\
            };
            
            \matrix[right = 1in of A5, chromosome] (B5)
            {
               4 \& 
              |[fill=white!80!blue]| \bf8 \& 
			   6 \& 
               9 \& 
               5 \& 
               3 \& 
              |[fill=white!80!black]|\bf 2 \& 
               1 \& 
              10 \& 
              -7 \\
            };

			\draw[->] (A5) -- node[above] {$L_5$} (B5) ;
		
            \matrix[chromosome, below = 1em of A5] (A6)
            {
               4 \& 
              |[fill=white!80!black]|-2 \& 
			   6 \& 
               9 \& 
               5 \& 
               3 \& 
              |[fill=white!80!blue]|-8 \& 
               1 \& 
              10 \& 
              -7 \\
            };
            
            \matrix[right = 1in of A6, chromosome] (B6)
            {
               4 \& 
              |[fill=white!80!blue]| \bf8 \& 
			   6 \& 
               9 \& 
               5 \& 
               3 \& 
              |[fill=white!80!black]| -2 \& 
               1 \& 
              10 \& 
              -7 \\
            };

			\draw[->] (A6) -- node[above] {$L_6$} (B6) ;
            \matrix[chromosome, below = 1em of A6] (A7)
            {
               4 \& 
              |[fill=white!80!black]|-2 \& 
			   6 \& 
               9 \& 
               5 \& 
              |[fill=white!80!blue]| 3 \& 
              |[fill=white!80!blue]|-8 \& 
              |[fill=white!80!blue]| 1 \& 
              10 \& 
              -7 \\
            };
            
            \matrix[right = 1in of A7, chromosome] (B7)
            {
               4 \& 
              6 \& 
	      9 \& 
               5 \& 
              |[fill=white!80!blue]| 3 \& 
              |[fill=white!80!blue]| \bf8 \& 
              |[fill=white!80!black]|-2 \& 
              |[fill=white!80!blue]| 1 \& 
              10 \& 
              -7 \\
            };

			\draw[->] (A7) -- node[above] {$L_7$} (B7) ;
        
        \end{tikzpicture}%
    }
    \caption{Illustration of $L_1$ to $L_7$. The boldface means the type of the node is converted.}
    \label{fig:neighborhood} 
\end{figure}
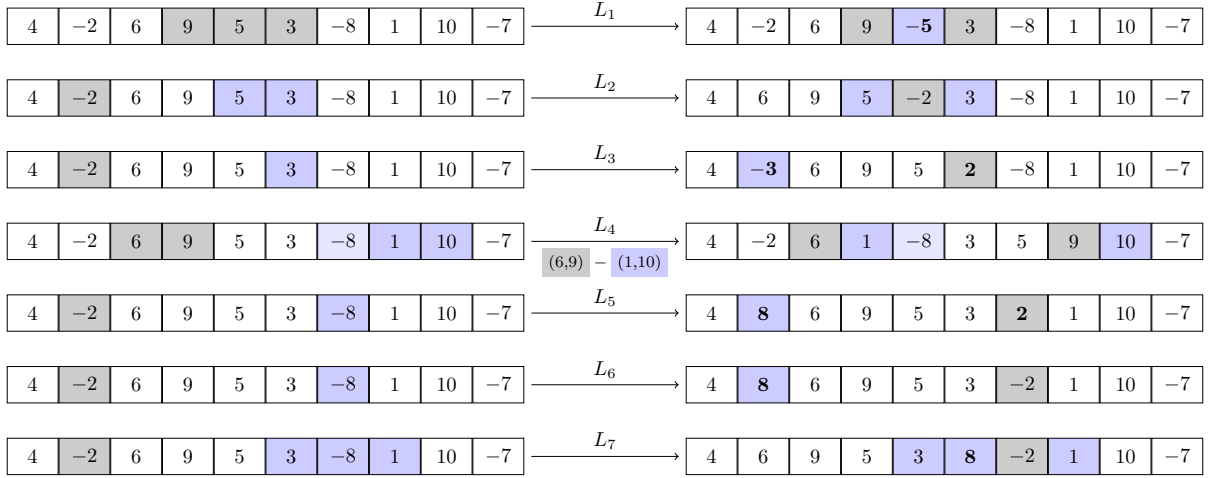

We introduce 7 new local search neighborhoods, labeled $L_1$ to $L_7$, which are shown to make further improvements.
A detailed description of these neighborhoods is provided below, and they are also illustrated in Figure \ref{fig:neighborhood}. 
A numerical analysis is performed in Section \ref{HaSubsection} in order to evaluate the contribution of the new local search neighborhoods.

\begin{itemize}
    \item $L_1$, (Convert to drone): Randomly choose three consecutive truck nodes and convert the middle one to a drone node. 
    \item $L_2$, (Relocate a drone node): Remove a random drone node and randomly locate it between two consecutive truck nodes as a drone node. 
    \item $L_3$, (Swap truck-drone nodes): Choose a truck node and a drone node randomly and swap them while keeping the type of each position.
    \item $L_4$, (Swap truck arcs): randomly select two arcs from the truck tour and swap them. The truck sequence as well as the drone sequence between the two arcs will be reversed. 
    \item $L_5$, (Swap drone nodes and convert both): Randomly choose two drone nodes and swap them while promoting their type to be truck nodes.
    \item $L_6$, (Swap drone nodes and randomly convert one): Same as $L_5$, only change the type of one node.
    \item $L_7$, (Insert a drone node): Randomly choose a drone node $d$ and a drone tuple $\langle i,j,k \rangle$, change $j$ to truck node and insert $d$ as either $\langle i,d,j,k \rangle$ or $\langle i,j,d,k \rangle$.
\end{itemize}

It is necessary to note that during the process of choosing the nodes randomly, we only select out of the $n_{\textsc{CLOSE}}$ nearest ones, instead of all nodes, in order to save time. 
As previously indicated, following the generation of an offspring, a thorough evaluation for feasibility ensues. 
In the event that the offspring is deemed infeasible, with a probability denoted as $P_{\textsc{repair}}$, a repair mechanism is invoked to rectify its infeasibility, thereby rendering it feasible. 
The resulting feasible offspring is subsequently integrated into the feasible subpopulation. 
Conversely, an infeasible offspring is directed to the corresponding infeasible subpopulation based on the type of its infeasibility.
For feasible offspring, an additional refinement step is introduced through the application of various local search neighborhoods, encompassing $N_1$ to $N_{15}$ and $L_1$ to $L_7$.
The procedure involves \bluenote{sequentially applying each neighborhood to the offspring once, until an improvement occurs}. 
Importantly, the order of neighborhood application is not fixed but rather undergoes random shuffling for every instance of offspring improvement. 
This approach ensures a diversified exploration of the solution space, contributing to the potential enhancement of the offspring's quality.

\subsection{Population Management} \label{pop}

A population management mechanism similar to that proposed in \citet{vidal2012hybrid} is used in this study with one exception. 
Their study incorporates two subpopulations, feasible and infeasible.
In our problem, there are three types of subpopulations: feasible, Type 1 infeasible (consecutive drone visits), and Type 2 infeasible (drone range violations). 
Subpopulations consist of at most $\mu + \lambda$ individuals, where $\mu$ denotes the minimum size of the subpopulation and $\lambda$ represents the offspring pool size. 
Subpopulations are first initialized with size $\mu$ according to Algorithm \ref{alg:init}. 
Each iteration will lead to the addition of the offspring to the corresponding subpopulation. 
Upon reaching the maximum size, $\lambda$ individuals will be eliminated through a survivors selection process in order to reduce the population size back to $\mu$. 
The top $\mu$ individuals are selected in accordance with makespan for feasible subpopulation and penalized makespan for infeasible subpopulations. 
Other important components of population management include penalty adjustment and diversification.

\paragraph{Penalty adjustment.} 
During each iteration, the penalty parameters can be dynamically modified in order to control the type of offspring generated.
The penalty parameters, $w_1$ and $w_2$, are used in the \JOIN{} algorithm and were defined in Section \ref{DP}.
We initialize the penalties with $w_1=2$ and $w_2=2$, and then update them during each iteration. 
Let $\xi^{REF}$ be the target proportion of the feasible subpopulation size to the entire population size.
In the case we have two infeasible subpopulations, their proportions will be maintained equally. 
Let $\xi^F$, $\xi^{M}$, and $\xi^{R}$ denote the proportion of feasible, Type 1 and Type 2 infeasible subpopulations in the last 100 individuals generated, similar to \citet{vidal2012hybrid}. 
The following adjustment will be performed in every iteration: 

\begin{itemize}
    \item if $\xi^F < \xi^{REF} - \zeta$
    \begin{itemize}
        \item if $\xi^R < \xi^{M}$, then $w_1 = \min\{\eta_I w_1,\ w_1^{\max}\}$
        \item else $w_2 = \min\{\eta_I w_2, \ w_2^{\max}\}$
    \end{itemize}
    \item if $\xi^F > \xi^{REF} + \zeta$
    \begin{itemize}
        \item if $\xi^R < \xi^{M}$, then $w_2 = \max\{\eta_D w_2,\ w_2^{\min}\}$
        \item else $w_1 = \max\{\eta_D w_1,\ w_1^{\min}\}$
    \end{itemize}
\end{itemize}
where $\zeta$ is the tolerance level and $\eta_I, \eta_D$ are multipliers to increase and decrease the penalty. 
We define $w_1^{\max}$ and $w_2^{\max}$ in order to prevent penalties from exploding, and $w_1^{\min}$ and $w_2^{\min}$ in order to prevent them from being smaller than 1. 

\paragraph{Diversification.} 
When no improvement is achieved after $It_{\textsc{DIV}}$ iterations, we employ a plan to enhance the population's genetic value. 
From each subpopulation, $n_{\textsc{BEST}}$ individuals will be retained based on fitness, while the rest will be discarded. 
The size of each subpopulation will then reach $\mu$ using the same strategy as the initial population algorithm.

\subsection{Escape Strategy} 
\label{sub:escape}

Typical meta-heuristic methods have the problem of solutions confined to a local optimal area, especially when local search is used, as in our algorithm. 
GAs offer a variety of tools for preventing premature convergence to local optima. 
Examples of such mechanisms include increasing population size and promoting diversification. 
It is noteworthy that while genetic variation can also be introduced through the process of mutation, this approach is inherently stochastic. 
Moreover, it is essential to underscore that the primary role of mutation in GAs is to prevent genetic drift rather than to solely address the premature convergence of the entire population.
The complexity of the fitness function in TSPD, however, necessitates that we explore more intelligent and powerful methods for escaping the local optima. 
There is a possibility that the algorithm may be trapped in a local optimum point when it does not show any improvement over a certain number of iterations.
The idea behind our trick is to generate a buffer of individuals with slightly poorer objective function values and then to randomly select one in each subsequent iteration and use local search to generate a better solution hoping it will escape the local optimum point. 
As our experiments demonstrated, this technique is very effective for escaping from local optima. 
Detailed information about this method can be found in Algorithm \ref{alg:escape}.
The algorithm commences by initializing an empty list of individuals with a maximum capacity of $n_{\textsc{buffer}}$, adding the current population's best individual to this buffer. 
Over the course of $max_{iter}$ iterations, the algorithm performs the following operations. 
During each iteration, it randomly selects one individual from the buffer, chooses a local search neighborhood randomly, and applies \bluenote{a local search with} the selected neighborhood to the chosen individual. 
If the modified individual surpasses the current best individual, it is added to the buffer, and the best individual is accordingly updated. 
Conversely, if the newly generated individual exhibits a larger makespan than the best individual, it is incorporated into the buffer only if the gap is smaller than a designated threshold denoted as $\epsilon_l$. 
When adding a new individual to the buffer (Lines 10 and 13), it is simply added when the buffer is not full, and replaces the worst individual when it is full.  
Upon completion of the iterations, all individuals in the buffer that outperform the current population's best individual are assimilated into the population.
In our numerical experiments, we set $n_{\textsc{buffer}}=40$, $\epsilon_l=0.05$ and $\max_{\text{iter}}=10000$.

\begin{algorithm}
\caption{Escaping from Local Optima} \label{alg:escape}
\begin{algorithmic}[1]
\State $\omega_\textsc{local} \gets$ $\texttt{best}(\Omega_\textsc{f})$  \Comment local optimum point
\State $\omega_\textsc{best} \gets \omega_\textsc{local}$
\State $\Omega_\textsc{Candidates} \gets \{\omega_\textsc{best} \}$   \Comment set of chromosomes
\While{number of iterations $<\max_{\text{iter}}$}
    \State Randomly select $\omega$ from $\Omega_\textsc{Candidates}$ 
    \State Randomly select $L(\cdot)$ from the set of available local search \bluenote{neighborhoods}
    \State $\omega' \gets L(\omega)$
    \If {$\omega' \notin \Omega_\textsc{Candidates}$}
        \If{$C(\omega')<C(\omega_\textsc{best})$}
            \State add $\omega'$ to $\Omega_\textsc{Candidates}$
            \State $\omega_\textsc{best} \gets \omega'$ 
        \ElsIf{$(C(\omega')-C(\omega_\textsc{best}))/C(\omega_\textsc{best})< \epsilon_l$} \Comment $\epsilon_l$ is a preset threshold.
            \State add $\omega'$ to $\Omega_\textsc{Candidates}$
        \EndIf
    \EndIf
\EndWhile
\ForEach{$\omega$ in $\Omega_\textsc{Candidates}$} 
    \If{$C(\omega)<C(\omega_\textsc{local})$}
       \State $\Omega_\textsc{f} \gets \Omega_\textsc{f} \cup \{\omega\}$
    \EndIf
\EndFor
\end{algorithmic}
\end{algorithm}

As previously stated, our proposed method is referred to as HGA-TAC. 
In instances where the escape strategy is applied, we denote the method as HGA-TAC$^+$. 
Throughout our computational experiments, we address all problem instances both with and without the use of the escape strategy. 
Consequently, the reported results encompass both HGA-TAC and HGA-TAC$^+$. 
In preparation for the subsequent section, we find it advantageous to provide a consolidated summary of all parameters employed in our approach across the paper. 
A comprehensive list of these parameters is presented in Table \ref{tab:parameters}.

\begin{table}[]
    \caption{A summarized list of all parameters used in this study. 
    The last four parameters are only used in HGA-TAC$^+$ }\label{tab:parameters}
    \begin{tabular}{l p{10.5cm} l}
        \toprule
        Parameter & Description & Tuned Values\\
        \midrule
    $\mu$                          & Minimum size of each subpopulation         & $15$                                        \\
    $\lambda$                      & Offspring pool size                             & $25$                                    \\
    $K_{\textsc{Tournament}}$                & Number of randomly chosen individuals in parent selection procedure      & $5$          \\
    $n_{\textsc{elite}}$                   & Parameter used in calculating fitness based on diversity factor               & $0.2  n_{\textsc{POPULATION}}$     \\
    $n_{\textsc{best}}$                     & Number of individuals retained during each implementation of diversification   & $0.3 \mu$  \\
    $P_{\textsc{repair}}$                   & Probability of repairing an infeasible generated offspring          & $0.5$               \\
    $P_M $                       & Probability of mutation                                               & $0.1$            \\
    $\zeta$                        & Tolerance level in penalty adjustment procedure                           & $0.05$        \\
    $\eta_I$                      & Penalty increasing multiplier                                      & $1.1$               \\
    $\eta_D$                     & Penalty decreasing multiplier                                        & $0.9$              \\
    $\xi^{REF}$                 & Target proportion of the feasible individual generation             & $0.2$              \\
    $w_1^{\max}, w_1^{\min}$                     & Maximum and minimum value allowed for penalty 1        &  $8$, $3$                          \\
    $w_2^{\max}, w_2^{\min}$                   & Maximum and minimum value allowed for penalty 2          & $5$, $1.5$                        \\
    $It_{\textsc{ni}}$                    & The number of iterations with no improvements for stopping the algorithm  &  $2500$       \\
    $It_{\textsc{div}}$                     & The number of iterations with no improvements for performing the diversification & $100$\\
    $It_{\textsc{loc}}$                     & The number of iterations with no improvements for performing the escape strategy & $1000$\\
    $n_{\textsc{buffer}}$                    & Maximum buffer size in escape strategy                                  & $40$           \\
    $\epsilon_l$           & Threshold for accepting the new solution in escape strategy                 & $0.05$       \\
    $\max_{\text{iter}}$                  & Maximum number of iterations in escape strategy           & $10000$    \\
    \bottomrule                    
    \end{tabular}
\end{table}

\section{Computational Results} \label{sec:Results}

In this section, our focus is on assessing the \bluenote{effectiveness} of our algorithm through experimentation on five widely recognized benchmark sets. 
First, we will outline the hardware and experimental settings as well as the parameters employed in the implementation of our algorithm and the methods applied for their refinement. 
Then, we will evaluate the effectiveness of HGA-TAC by comparing it to optimal solutions using randomly generated instances. 
Subsequently, we will provide a brief overview of the benchmark sets utilized for comparative analysis. 
A comprehensive summary of all results will be presented in \bluenote{Tables} \ref{tab:summary} and \ref{tab:count}. 
Following this, we will delve into a detailed examination of our algorithm's performance on each benchmark set individually. 
For a more granular breakdown, please refer to the Appendix, where \bluenote{Tables} \ref{tab:Agatz1Alpha1}-\ref{tab:Ha} present the detailed results for each benchmark set.

\subsection{Experimental conditions}
The algorithm is coded and implemented in Julia programming language and is executed on a Mac computer with 16 GB of RAM and an Apple M1 processor. 
In the process of parameter tuning, we meticulously refined a significant portion of the parameters throughout the developmental phase of the algorithm. 
The refinement was a gradual and iterative process, closely aligned with the ongoing enhancements in the codebase. 
Specifically, for parameters $\mu, \lambda, n_{\textsc{CLOSE}}$, and $It_{\textsc{DIV}}$, we employed a grid search methodology. 
This involved systematically exploring various combinations of these parameters using random samples extracted from diverse benchmark sets. 
The ultimate selection of parameter combinations was based on achieving the best results on average across these experiments.
The values of all parameters utilized in this study after tuning are documented in Table \ref{tab:parameters}.

Given the stochastic nature of GA, the outcomes may vary across individual runs on a specific instance. 
To ensure a comprehensive evaluation and equitable comparison with alternative methodologies, it is common practice to execute the algorithm multiple times and analyze the aggregated performance metrics. 
In this study, each instance is subjected to $10$ independent runs, and the reported results encompass both the best and average outcomes derived from these $10$ executions, respectively denoted as ``Best'' and ``Avg.'' in the tables.

\subsection{Comparative Analysis with Exact Solutions}
In preparation for evaluating our algorithm on established benchmark datasets, we conducted a preliminary examination of HGA-TAC's accuracy on small-sized instances, where obtaining optimal solutions is feasible. 
To achieve this, we randomly generated small TSPD instances and solved them using the MILP formulation proposed by \citet{roberti2021exact}, implemented with Gurobi 9.5. 
The results were subsequently compared to those obtained by HGA-TAC. 
The random instance generation process involved selecting the depot location from a uniform distribution $[0,1] \times [0,1]$, while customer locations were chosen from a uniform distribution $[0,10] \times [0,10]$. 
Instances with 11, 12, and 13 customers were generated, with 10 instances for each size. 
These instances are publicly accessible in our GitHub repository.

Each of the 30 instances is solved with $\alpha$ values of 1, 2, and 3, representing the ratio of the drone's speed to the truck's speed, assuming an unlimited drone flying range. 
The computational times, measured in seconds, were recorded, and a time limit of 3600 seconds was set for MILP solutions. 
Table \ref{tab:exactGap}, \bluenote{which can be found in the appendix,} summarizes the results for these 90 instances. 
The columns ``Gap$^b$'' and ``Gap$^a$'' denote the gap between the ``Best'' and ``Avg.'' results of HGA-TAC compared to the optimal solutions. 
It is noteworthy that HGA-TAC's ``Best'' results were nearly equivalent in almost all instances (88 out of 90). 

\subsection{Benchmark sets}
Our algorithm, tailored for addressing the TSPD and the FSTSP, undergoes rigorous evaluation across five meticulously chosen benchmark sets. 
These sets serve the dual purpose of assessing both TSPD and FSTSP instances, allowing for a comprehensive evaluation against state-of-the-art algorithms documented in the literature. 
The nomenclature of the benchmark sets in this study adheres to a systematic convention: The term 'Set' is followed by the initial character of the respective first author's name from whose work the sets are derived. 
The benchmark sets encompass a variety of scenarios, each strategically selected to enable robust comparisons. 
The TSPD benchmark sets utilize the Euclidean distance metric for both trucks and drones, while the FSTSP benchmark sets employ Manhattan and Euclidean metrics for trucks and drones, respectively. \bluenote{We consider the following benchmarks:}

\begin{itemize}
\item \textbf{Set A$_u$}: Developed by \citet{agatz2018optimization}, this set spans instances ranging from $10$ to $250$ nodes, designed to test TSPD with an unlimited drone flying range. 
Instances are categorized based on \emph{uniform}, \emph{single-center}, or \emph{double-center} distributions. 
In uniform instances, the $x$ and $y$ coordinates for each location are independently and uniformly drawn from the set $\{1, 2, ..., 100\}$, resulting in instances where the distribution of points is evenly spread across the available coordinate space.
For single-center instances, each location follows a different approach. 
An angle $a$ is uniformly drawn from the interval $[0, 2\pi]$, and a distance $r$ is drawn from a normal distribution with a mean of $0$ and a standard deviation of $50$. 
The $x$ coordinate of the location is then calculated as $r cos a$, and the $y$ coordinate is calculated as $r sin a$. 
This methodology introduces a circular clustering effect, with locations concentrated around a central point.
Double-center instances build upon the single-center concept. 
Similar to the single-center instances, locations are generated with an angle and distance. 
However, in this case, every location has a $50\%$ chance of being translated by $200$ distance units along the $x$-axis. 
This introduces a bimodal distribution, creating instances where locations are clustered around two distinct centers, adding an additional layer of complexity to the spatial arrangement.
Each instance in this benchmark set is evaluated for three different values of $\alpha$ (the ratio of drone speed to truck speed: $\alpha \in \{1, 2, 3\}$). 
The set comprises $630$ instances, and baseline algorithms for comparison include DPS$/25$ \citep{bogyrbayeva2023deep} and HGVNS \citep{de2020variable}.

\item \textbf{Set A$_l$}: Also introduced by \citet{agatz2018optimization}, this set focuses on TSPD with a limited drone flying range. 
Instances range from $10$ to $100$ nodes, with variations in drone range for each size. 
The uniform distribution is used for generating these instances.
A total of $320$ instances are included, and baseline algorithms for comparison are TSP-ep-all \citep{agatz2018optimization} and DPS$/25$ \citep{bogyrbayeva2023deep}.

\item \textbf{Set B}: Created by \citet{bogyrbayeva2023deep}, this set consists of two subsets of TSPD instances with an unlimited drone range for $\alpha = 2$. 
The subsets, 'Random' and 'Amsterdam' differ in the method of instance generation. 
In the first subset of instances, a uniform distribution over $[0,1]\times[0,1]$ and $[0,100]\times [0,100]$ is used to sample the $x$ and $y$ coordinates of the depot and customer nodes, respectively. 
This makes the depot to be always located in the lower left corner. 
The method of generation is similar to that proposed by \citet{agatz2018optimization}. 
For each size of $20, 50$, and $100$ nodes, $100$ samples of instances are presented. 
The second subset of instances includes $100$ samples for each size of $10, 20$, and $50$ nodes. 
The depot in this subset of instances, is randomly chosen out of the nodes. 
This benchmark set is chosen for its relevance in comparing our algorithm against the best existing Deep Reinforcement Learning (DRL) method for TSPD. 
Baseline algorithms include TSP-ep-all \citep{agatz2018optimization}, DPS$/25$, and the DRL-based Hybrid Model (HM) \citep{bogyrbayeva2023deep}.

\item \textbf{Set M}: Generated by \citet{murray2015flying}, this FSTSP benchmark set comprises $36$ instances with nodes distributed over an $8$ miles by $8$ miles area. 
Each instance includes $10$ customers, of which $8$ or $9$ are eligible for drone visits. 
Using the Manhattan metric, the truck is assumed to be traveling at $25$ miles per hour, and using the Euclidean distance metric, the drone is assumed to be traveling at $15, 25$, or $35$ miles per hour. 
Both the launch and retreat times, $s_L$ and $s_T$, are set to one minute. 
The drone's endurance is either $20$ or $40$ minutes, which results in a total of $72$ instances.
Baseline algorithms include \citet{murray2015flying}'s heuristic approach and HGA20 \citep{ha2020hybrid}.

\item \textbf{Set H}: Introduced by \citet{ha2020hybrid}, this set includes $60$ instances with $10, 50$, and $100$ customers. 
Assumptions align with FSTSP by \citet{murray2015flying}. 
The benchmark is designed for comparison with the HGA20 algorithm. 
Parameters include Manhattan distance for trucks, Euclidean distance for drones, and specific operational characteristics for both.
Distance matrices are calculated using Manhattan distance for trucks and Euclidean distance for drones. 
The drone and truck have both been set to operate at a speed of $40$ km/h. 
The drone is designed to fly for $20$ minutes at a time.
Launch time $S_L$ and retrieval time $S_R$ are both $1$ minute and only $80\%$ of the customers are eligible to be serviced by drones.
This benchmark set is chosen to compare the performance of our algorithm with HGA20 \citep{ha2020hybrid}.

\end{itemize}
A comprehensive overview of the features of these benchmark sets is provided in Table \ref{tab:benchmarks}.

\begin{table}[] \centering
    \caption{A summary of the benchmark sets used in this study.}
    \label{tab:benchmarks}
    \begin{tabular}{ c l l l l }
        \toprule
    Benchmark Set & Problem & Size Range & Source Paper & Baseline Algorithms \\
    \midrule
    Set A$_u$                           & TSPD                      & $10-250 $                             & \citet{agatz2018optimization}                   & HGVNS, DPS$/25$                              \\
    Set A$_l$                           & TSPD                             & $10-100 $                             & \citet{agatz2018optimization}                   & TSP-ep-all, DPS$/25$                         \\
    Set B                               & TSPD                             & $10-100$                              & \citet{bogyrbayeva2023deep}                     & TSP-ep-all, DPS$/25$, HM                     \\
    Set M                               & FSTSP                            & $10 $                                 & \citet{murray2015flying}                        & MC, HGA20                               \\
    Set H                               & FSTSP                            & $10-100 $                             & \citet{ha2018min}                               & HGA20                                  \\
    \bottomrule
    \end{tabular}
\end{table}

\begin{table}[]
    \caption{A summary of results for HGA-TAC and HGA-TAC$^+$ for all sets of instances.
      Time* means the time as reported in \citet{de2020variable} for HGVNS, \citet{ha2020hybrid} for HGA20 and \citet{bogyrbayeva2023deep} for HM (4800). Computational times are measured in seconds.}
    \label{tab:summary}
    \begin{adjustbox}{max width=\textwidth}
        \begin{tabular}{l r rr rr rr c}
            \toprule
                & \multicolumn{1}{c}{Baseline} 
                & \multicolumn{2}{c}{DPS/25}                 
                & \multicolumn{2}{c}{HGA-TAC} 
                & \multicolumn{2}{c}{HGA-TAC$^+$} 
				& Details \\
            \cmidrule(lr){2-2} \cmidrule(lr){3-4} \cmidrule(lr){5-6} \cmidrule(lr){7-8} 
            Instance set 
                & Time*
                & Gap & Time
                & Gap & Time
                & Gap & Time
                	& (Section)
                \\
            \midrule
            \bf TSPD            & \multicolumn{1}{c}{HGVNS} \\
            \cmidrule(lr){2-2}
            Set A$_u$ ($\alpha=1$) 	&  43.45 & -10.45\% &  2.51 & -10.66\% &  6.00 & \bf -11.14\% & 34.06   & Table \ref{tab:Agatz1Alpha1} (\S \ref{Agatz1Subsection}) \\
            Set A$_u$ ($\alpha=2$) 	&  41.16 & -3.71\% &  4.78 & -3.17\%  &  8.32 & \bf -4.57\%   & 45.98   & Table \ref{tab:Agatz1Alpha2} (\S \ref{Agatz1Subsection}) \\
            Set A$_u$ ($\alpha=3$) 	 &  41.19 & -1.20\% &  6.56 & -2.07\%  & 10.33 & \bf -4.15\% & 61.44   & Table \ref{tab:Agatz1Alpha3} (\S \ref{Agatz1Subsection}) \\
            Set A$_l$          		&          - &  0.00\% &  0.50 & 0.07\% &  2.23 & \bf -0.52\% &  8.12   & Table \ref{tab:Agatz2} (\S \ref{Agatz2Subsection}) \\
            \midrule
            \bf TSPD            & \multicolumn{1}{c}{HM (4800)} \\
            \cmidrule(lr){2-2}
            Set B (rand)   &         2.07 & 1.02\% &  0.56 & 0.87\%  &  2.18 & \bf -0.53\% &  8.16   & Table \ref{tab:Aigerim} (\S \ref{AigerimSubsection}) \\
            Set B (Ams)    &        0.75 & 0.02\% &  0.22 & -0.67\% &  0.69 & \bf -1.30\% &  2.29   & Table \ref{tab:Aigerim} (\S \ref{AigerimSubsection}) \\
			\midrule
            \bf FSTSP 			& \multicolumn{1}{c}{HGA20} \\
            \cmidrule(lr){2-2}
            Set M              &     -  &      - &     - & -0.28\% &  0.40 & \bf -0.48\% &  1.16   & Table \ref{tab:Murray} (\S \ref{MurraySubsection}) \\ 
            Set H  		   & 159.60 &      - &     - & -0.47\% &  3.60 & \bf -1.13\% & 16.20   & Table \ref{tab:Ha}  (\S \ref{HaSubsection})\\

            \bottomrule
        \end{tabular}
    \end{adjustbox}
\end{table} 

\subsection{Summarized results}
In Table \ref{tab:summary}, a concise summary of results is provided for all sets of instances, offering a high-level overview of the algorithmic performance. 
However, for a detailed examination of the outcomes for each set of instances, the reader is directed to Tables \ref{tab:Agatz1Alpha1}--\ref{tab:Ha} in the Appendix. 
It is noteworthy that TSP-ep-all is exclusively utilized for sets A$_u$ and B; therefore, its results are not included in this summary table. 
Instead, interested readers can refer to Tables \ref{tab:Agatz2} and \ref{tab:Aigerim} in the Appendix for a comprehensive analysis of TSP-ep-all outcomes. 
In the context of TSPD, sets A$_u$ and A$_l$ are segregated from set B due to differences in the primary baseline algorithm; for set A$_u$ the primary baseline is HGVNS, whereas for set B the primary baseline is HM (4800). 
Furthermore, for the FSTSP results, DPS$/25$ is not applicable, as it is specifically designed for solving TSPD. 
\bluenote{The ``Gap'' is determined based on the average results obtained over 10 independent runs.}
Additionally, computational times for HGA20 on set M were not reported in \citet{ha2020hybrid}, leading to the absence of time-related information in Table \ref{tab:summary}.
A cursory examination of the HGA-TAC$^+$ column reveals competitive performance with acceptable computational times among the existing algorithms in the table.

\begin{table}[] \centering
    \caption{Number of instances that HGA-TAC and HGA-TAC$^+$ found the best solution for all sets of instances.}
    \label{tab:count}
    \begin{adjustbox}{max width=\textwidth}
	\begin{tabular}{lrrrrrrrrrrrr}
	\toprule
                                 & \multicolumn{1}{c}{}      & \multicolumn{5}{c}{Better or equal}                                                                                              & \multicolumn{1}{c}{} & \multicolumn{5}{c}{Strictly better}                                                                                              \\
                                 \cmidrule{3-7}         \cmidrule{9-13}
                                 & \multicolumn{1}{c}{}      & \multicolumn{2}{c}{HGA-TAC}                         & \multicolumn{1}{c}{} & \multicolumn{2}{c}{HGA-TAC$^+$}                     & \multicolumn{1}{c}{} & \multicolumn{2}{c}{HGA-TAC}                         & \multicolumn{1}{c}{} & \multicolumn{2}{c}{HGA-TAC$^+$}                     \\
                                         \cmidrule{3-4}          \cmidrule{6-7}           \cmidrule{9-10}        \cmidrule{12-13}
                              Instance set   & \multicolumn{1}{c}{Total} & \multicolumn{1}{c}{Best} & \multicolumn{1}{c}{Avg.} & \multicolumn{1}{c}{} & \multicolumn{1}{c}{Best} & \multicolumn{1}{c}{Avg.} & \multicolumn{1}{c}{} & \multicolumn{1}{c}{Best} & \multicolumn{1}{c}{Avg.} & \multicolumn{1}{c}{} & \multicolumn{1}{c}{Best} & \multicolumn{1}{c}{Avg.} \\
                                 \midrule
\textbf{TSPD} &&&&&&&&&&&\\
Set A$_u$ ($\alpha=1$)               & 210                       & 160                      & 124                      &                      & 202                      & 164                      &                      & 114                      & 82                       &                      & 153                      & 120                      \\
Set A$_u$ ($\alpha=2$)               & 210                       & 144                      & 94                       &                      & 193                      & 148                      &                      & 122                      & 78                       &                      & 171                      & 132                      \\
Set A$_u$ ($\alpha=3$)               & 210                       & 186                      & 130                      &                      & 209                      & 184                      &                      & 165                      & 116                      &                      & 188                      & 169                      \\
Set A$_l$                  & 320                       & 222                      & 163                      &                      & 262                      & 202                      &                      & 75                       & 47                       &                      & 115                      & 74                       \\
Set B                            & 600                       & 426                      & 250                      &                      & 536                      & 361                      &                      & 308                      & 183                      &                      & 423                      & 284                      \\
\midrule
\textbf{TSPD Total}              & \textbf{1550}             & \textbf{1138}            & \textbf{761}             & \textbf{}            & \textbf{1402}            & \textbf{1059}            & \textbf{}            & \textbf{784}             & \textbf{506}             & \textbf{}            & \textbf{1050}            & \textbf{779}             \\
\midrule
\textbf{FSTSP} &&&&&&&&&&&\\
Set M                            & 72                        & 64                       & 58                       &                      & 67                       & 61                       &                      & 29                       & 29                       &                      & 30                       & 29                       \\
Set H                            & 60                        & 38                       & 33                       &                      & 44                       & 39                       &                      & 38                       & 33                       &                      & 44                       & 39                       \\
\midrule
\textbf{FSTSP Total}             & \textbf{132}              & \textbf{102}             & \textbf{91}              & \textbf{}            & \textbf{111}             & \textbf{100}             & \textbf{}            & \textbf{67}              & \textbf{62}              & \textbf{}            & \textbf{74}              & \textbf{68}              \\
\midrule
\textbf{TOTAL}                   & \textbf{1682}             & \textbf{1240}            & \textbf{852}             & \textbf{}            & \textbf{1513}            & \textbf{1159}            & \textbf{}            & \textbf{851}             & \textbf{568}             & \textbf{}            & \textbf{1124}            & \textbf{847}       \\
\bottomrule     
	\end{tabular}
    \end{adjustbox}
\end{table}

Not only does our algorithm perform the best among available algorithms in the average sense, but it also finds the new best solutions for many benchmark instances.
We report the number of instances in which our algorithms found a solution that was either equal to or better than the best existing solution. 
The results for each set of instances can be found in Table \ref{tab:count}. 
\bluenote{The columns labeled ``Best'' and ``Avg.'' denote the number of instances where our algorithms' best and average results (over 10 runs) are comparable with the best-know solutions, respectively. 
For example, in the last column of the first, ``$120$'' indicates that HGA-TAC$^+$'s average results over 10 run are better than best-known solutions in $120$ out of $210$ instances.}
The detailed results for set A$_u$ are compared only with those obtained by DPS$/25$. 
Thus, we cannot claim that our algorithms have found the best solutions, since running TSP-ep-all was not feasible due to the large problem sizes, and we did not have access to the detailed HGVNS results.
 Nevertheless, we have improved solutions in $512$ out of $630$ instances in comparison with DPS/25. 
Among the $320$ examples of set A$_l$, our algorithms have demonstrated better or similar performance in $262$ of them and strictly better performance in $115$ of them, \bluenote{in terms of objective function}. 
Based on $600$ instances in set B our algorithms have found better or equal solutions in $536$ instances and strictly \bluenote{better} solutions in $423$ instances, \bluenote{in terms of objective function}.
Excluding Set A$_u$, where results were compared only with DPS/25, a total of $920$ TSPD instances from Sets A$_l$ and B, where TSP-ep-all solutions were available, showed that our algorithms improved upon the best existing solutions in $538$ instances.
A similar performance can be observed for our algorithms when solving instances of FSTSP. 
Out of $72$ instances of set M, we have found better or similar solutions in $67$ instances and strictly better solutions in $30$ instances. 
Out of $60$ instances, $44$ were improved by our algorithms for set H. 
As a result, we have found better solutions than the existing best for $74$ of the $132$ instances of FSTSP.

\subsection{Detailed results}

\subsection*{Results for the TSPD Instances with Unlimited Flying Range from \citet{agatz2018optimization}}  \label{Agatz1Subsection}

In this section, we present a comprehensive analysis of the experiments conducted on set A$_u$, with detailed results provided in Tables \ref{tab:Agatz1Alpha1}, \ref{tab:Agatz1Alpha2}, and \ref{tab:Agatz1Alpha3} for $\alpha=$ 1, 2, and 3, respectively. 
These tables are available in the Appendix.
To ensure a fair comparison, we have coded the DPS/25 in Julia and run it on the same machine. 
It is important to note, however, that the HGVNS method suggested by \citet{de2020variable} was implemented in C++ and executed on a Core i7 processor with 3.6 GHz along with 16 GB of RAM.
According to \url{www.cpubenchmark.net}, our processor is 2.52 times faster than theirs. 
HGVNS has an average run time of 41.93 seconds; however, after converting it to our processor performance, it will be 16.64 seconds, while HGA-TAC, HGA-TAC$^+$, and DPS/25 have run times of 8.22, 47.16, and 1.11 seconds respectively. 
HGA-TAC is faster than HGVNS but slower than DPS/25. 
HGA-TAC has an average advantage of 5.30\% over HGVNS and a 0.17\% advantage over DPS/25. 
Where $\alpha=2,3$, HGA-TAC performs better than DPS/25, while DPS/25 provides better solutions when $\alpha=1$. 
In spite of being slower, HGA-TAC$^+$ shows an improvement of 6.62\% and 1.51\% over HGVNS and DPS/25, respectively. 
In all scenarios, HGA-TAC$^+$ produces better solutions than DPS/25.

\subsection*{Results for the TSPD Instances with Limited Flying Ranges from \citet{agatz2018optimization}}
\label{Agatz2Subsection}

The instances in set A$_l$ are based on TSPD with restricted drone ranges. 
The results are summarized in Table \ref{tab:Agatz2} which can be found in the Appendix.
The number of nodes in the instance is indicated in \bluenote{column $N$}, and the ratio of drone range to the maximum distance between a pair of locations as a percentage is indicated in \bluenote{column $r$}. 
Due to the fact that each drone operation involves traveling between two pairs of nodes, $r=200$ indicates an unlimited flying range. 
We examine the \bluenote{effectiveness} of our algorithm in this part by comparing the results with those of TSP-ep-all and DPS/25. 
In order to compare running times, all algorithms were developed in the same programming language (Julia) and executed on the same machine. 
As depicted in Table \ref{tab:Agatz2}, HGA-TAC has better performance in \bluenote{instances with} $n=10$ and $n=20$. 
In the case of $n=50$, HGA-TAC is still outperforming DPS/25, \bluenote{in terms of solution quality}.
This is not the case in the rest of the sizes. 
HGA-TAC$^+$, however, \bluenote{better objective function than} DPS/25 in all cases. 
As compared to TSP-ep-all, averaged over all instances, HGA-TAC$^+$ results in a 0.18\% gap, where the average running time is 7.02 seconds, compared to 29.84 seconds for TSP-ep-all.

\subsection*{Results for the TSPD Instances from \citet{bogyrbayeva2023deep}}
\label{AigerimSubsection}

In this section, we present a comprehensive analysis of the experiments conducted on set B, with detailed results provided in Table \ref{tab:Aigerim} which can be found in the Appendix.
The results of our HGA-TAC and HGA-TAC$^+$ algorithms are compared to those obtained with TSP-ep-all, DPS/25, and HM (4800) in \citet{bogyrbayeva2023deep}. 
HM stands for Hybrid Model based on deep reinforcement learning, and the HM (4800) represents the best solution generated by the neural network from 4800 samples. 
HGA-TAC and HGA-TAC$^+$ have been run 10 times on each instance and we report the average and best results.
The computational time for HGA-TAC(Best) and HGA-TAC$^+$(Best) are obtainable by multiplying the corresponding run time by 10. 
TSP-ep-all and DPS/25 are implemented in our machine for a fair comparison of running time, while we present the time for HM (4800) reported in \citet{bogyrbayeva2023deep}, implemented on NVIDIA A100 GPU (80 GiB) and AMD EPYC 7713 64-Core Processor CPU (128 threads used). 
Note, however, that training time is not included in the reported times for HM (4800). 
For problems with a size smaller than 25, DPS/25 is identical to TSP-ep-all. 
Gaps are calculated in comparison with the results of TSP-ep-all as $\dfrac{(z - \bar{z})}{\bar{z}}\times 100$, where $z$ and $\bar{z}$ are the costs of each algorithm and TSP-ep-all respectively. 
For HGA-TAC and HGA-TAC$^+$, the gaps are calculated based on the average values. 
Negative values for the gap indicate that the algorithm is outperforming TSP-ep-all.
Among the algorithms, HGA-TAC$^+$(Best) appears to perform the best on all instance sizes in both datasets.
Compared to TSP-ep-all, HGA-TAC$^+$(Avg.) shows better results in four of six cases and is slightly better overall.
With instances of size 20, all of our algorithms outperform all other methods, \bluenote{in terms of solution quality}. 
For larger sizes, however, the algorithms are in competition with one another over different cases.

\subsection*{Results for FSTSP instances from \citet{murray2015flying}}
\label{MurraySubsection}
The next step is to test our algorithm using the FSTSP configuration.
The results of our experiments on set M instances can be found in Table \ref{tab:Murray} in the Appendix.
Using this collection of examples, we compare the performance of our method to that proposed by \citet{murray2015flying} as well as the HGA20 in \citet{ha2020hybrid}.
In the same manner as HGA20, we solve each instance ten times and report both the best and average results. 
According to Table \ref{tab:Murray}, our algorithms, HGA-TAC and HGA-TAC$^+$, demonstrate significant improvements over existing methods, \bluenote{with respect to solution quality}. 
There is no mention of computational times in either \citet{murray2015flying} or \citet{ha2020hybrid}. 
However, we provide the run times so that future comparisons can be made.
In the column ``Gap'', we indicate the difference between the average results of our method and those of HGA20. 
It is, however, the best of MC and HGA20-Best that we use to count the instances in which HGA-TAC and HGA-TAC+ have improved the existing solutions. 
As a result, HGA-TAC found better solutions in 29 instances and similar solutions in 35 instances. 
There were only 8 instances where it lost to the best baseline algorithm. 
Those numbers are 30, 37, and 5 for HGA-TAC+, respectively. 
Compared to HGA20, HGA-TAC and HGA-TAC+ show average performance improvements of 0.28\% and 0.48\%, respectively.

\subsection*{Results for FSTSP instances from \citet{ha2018min}}
\label{HaSubsection} 
The results of our HGA-TAC and HGA-TAC$^+$ on set H instances are compared in Table \ref{tab:Ha} (see Appendix) with HGA20 in \citet{ha2020hybrid}, which is implemented in C++ and executed on a desktop computer with an Intel Core i7-6700, 3.4 GHz processor. 
Based on the same scaling system in Section \ref{Agatz1Subsection}, their running times are comparable with ours if divided by 1.82. 
Each instance is solved 10 times, and the best and average solutions with the running time are reported. 
In the referenced paper, the times are reported in minutes, but we report them here in seconds.
By scaling to our processor's performance, the average running time for HGA20 is 87.60 seconds against 3.73 and 16.29 seconds for HGA-TAC and HGA-TAC$^+$, respectively. 
HGA-TAC and HGA-TAC$^+$, having 0.47\% and 1.13\% improvement over HGA20 solutions and being significantly faster, are clearly delivering superior results. 

Moreover, the performance of $L_1$--$L_7$ local search \bluenote{neighborhoods} proposed in Section \ref{improve} has been evaluated by solving this set of problems with and without these neighborhoods. 
The results are presented in Table \ref{tab:sensitivity}, \bluenote{in which the columns labeled ``Best" and ``Avg." are obtained by taking an average over the makespan of all instances.} The results indicate 0.45\% and 0.81\% improvement \bluenote{on average} caused by our proposed neighborhoods in HGA-TAC and HGA-TAC$^+$, respectively. \bluenote{We have included the average computational time for a fair comparison. Additionally, the column labeled ``Improved'' indicates the number of instances in which our results outperform those of HGA20. Notably, the introduction of new local search neighborhoods has contributed to the discovery of superior solutions in numerous instances within the dataset.}

\begin{table}\centering
    \caption{Contributions of new local search \bluenote{neighborhoods} $L_1$--$L_7$ }
    \label{tab:sensitivity}

\bluenote{
 \begin{tabular}{l rrrr rrrr}
        \toprule
              FSTSP Set H                  & \multicolumn{4}{c}{without $L_1$--$L_7$}  & \multicolumn{4}{c}{with $L_1$--$L_7$}  \\
                               \cmidrule(lr){2-5}  \cmidrule(lr){6-9}  
                                & Best        & Avg.     & Time  & Improved            & Best             & Avg.      & Time  & Improved        \\
                                \midrule
        HGA-TAC            & 261.66           & 263.55     & 3.13   & 31     & 260.66           & 262.37    &3.37 &  38  \\
        HGA-TAC$^+$   & 260.74            & 262.92     & 14.65  & 37       & 258.92           & 260.85   & 16.29 & 44                 \\
        \bottomrule
    \end{tabular}
}

\end{table}

\subsection{\bluenote{The effectiveness of} the Escape Strategy}
\label{Escape effectiveness}

\begin{figure}
    \centering
    \includegraphics[scale=0.7]{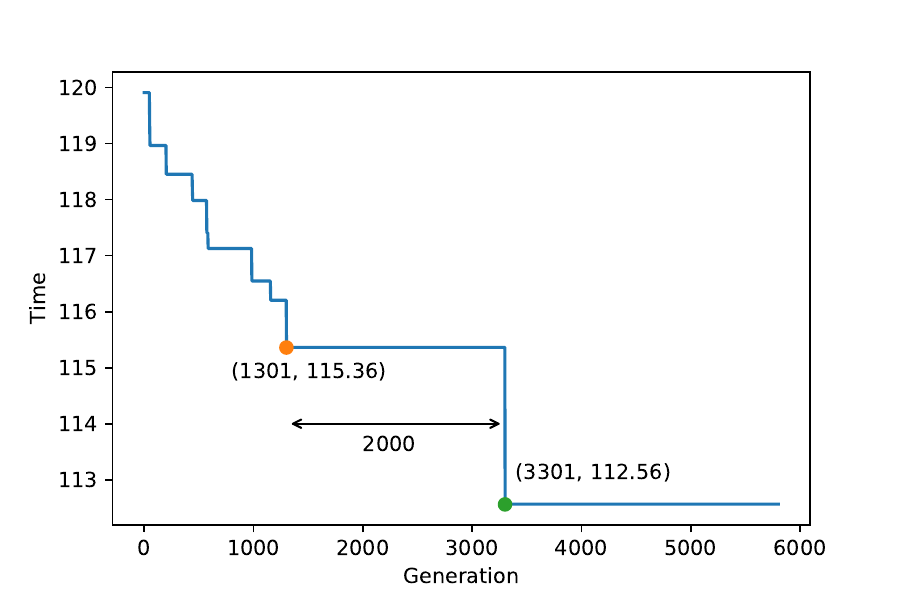}
    \caption{The objective function of instance B6 over generations, solved by HGA-TAC$^+$.
    The final solution found by HGA-TAC$^+$ is 112.56.}
    \label{fig:escape}
\end{figure}

This section aims to illustrate the effectiveness of the escape local optima plan. 
As stated previously, every 1000 iterations without any improvement, Algorithm \ref{alg:escape} is triggered and attempts to escape the local optimum point. 
A detailed representation of the instance B6 chosen from FSTSP Set H, as solved by HGA-TAC$^+$, is shown in Figure \ref{fig:escape}. 
After generation 1301, the GA appears to be trapped in the local optima with an objective value of 115.36. 
However, at the second attempt to escape the local optima, the algorithm manages to obtain an objective value of 112.56. 
This was an example of what happens behind the scenes of the HGA-TAC$^+$ algorithm compared to the regular HGA-TAC algorithm.

\begin{figure}
    \centering
    \includegraphics[width=\textwidth]{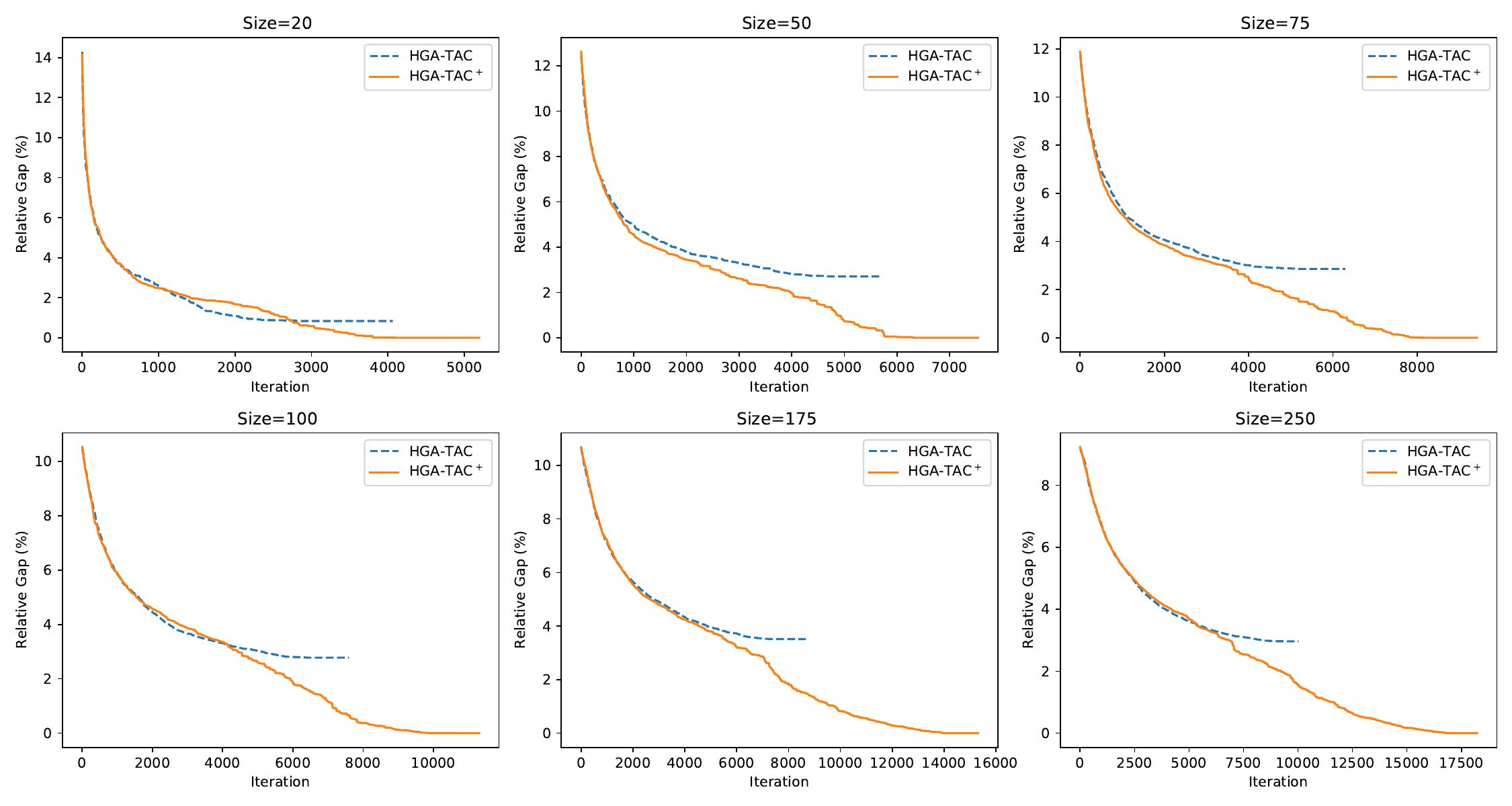}
    \caption{\bluenote{Evaluating the effectiveness of the Escape Strategy using benchmark set A$_u$.}}
    \label{fig:escape_effect}
\end{figure}

\bluenote{
To assess the effectiveness of the Escape Strategy, we employed instances from Set A$_u$. 
For this analysis, we solved instances of sizes 20, 50, 75, 100, 175, and 250, employing both HGA-TAC and HGA-TAC$^+$. 
Each instance size, represented as a graph in Figure \ref{fig:escape_effect}, encompasses 90 instances derived from three distributions and three alpha values, with 10 instances per setting.
To gauge the convergence behavior, we calculated the average relative gap over these instances. 
Initially, we standardized the iteration counts across all instances by interpolating the data and dividing the values by the best objective value found by HGA-TAC$^+$, ensuring uniformity in the number of iterations and a comparable scale among different instances. 
Subsequently, we computed the average gap across the instances within each group.
Upon analysis of Figure \ref{fig:escape_effect}, it is evident that the Escape Strategy exhibits minimal impact on instances of size 20. 
This observation suggests that HGA-TAC already attains near-optimal solutions for smaller instances, leaving limited room for further improvement. 
However, as the instance sizes increase, the Escape Strategy demonstrates a pronounced effect, notably enhancing the Makespan at the expense of additional iterations.
}

\section{Conclusions} \label{sec:Con}

This paper presented a Hybrid Genetic Algorithm with Type-Aware Chromosome encoding (HGA-TAC), to solve TSPD and FSTSP. 
The GA is responsible for partitioning and managing the sequences of the customers, while the DP is responsible for determining the rendezvous points and calculating the objective value. 
TSPD and FSTSP both aim to find the best routes for trucks and drones, yet several assumptions differ slightly between the two. 
There were several local search \bluenote{neighborhoods} offered, along with two crossover approaches that were specifically designed for our problem. 

A key innovation presented in this work is the ``escape strategy,'' detailed in Section \ref{sub:escape}, which effectively addresses the challenge of local optima inherent in meta-heuristic methods.
The ``escape strategy'' serves as a versatile tool for diversification within GAs, transcending the confines of our specific problem. 
By generating a buffer of individuals and strategically applying local search, the algorithm escapes local optima, contributing to improved solution quality. 
This strategy, outlined in Algorithm \ref{alg:escape}, has proven effective across various scenarios, as demonstrated in our numerical experiments. 
The adaptability of the ``escape strategy'' suggests its potential as a general diversification method for Genetic Algorithms, offering a promising avenue for future research.

As evidenced by our testing on five benchmark sets of instances, our GA either exceeded or was competitive with the best existing methods. 
We believe the faster performance in our method is due to making the decisions about the sequence and node types by GA and leaving less information to be decided optimally by DP. 
Moreover, it was observed that Escape \bluenote{Strategy} improved the quality of the solution at the cost of increased running time. 

A few future research directions may be considered.
A generalized version of the proposed method can be implemented using multiple drones in conjunction with a truck in order to solve the TSPD. 
A further extension could involve allowing the drone to visit more than one customer in a single operation. 
Furthermore, the proposed method in this paper could be extended to vehicle routing problems with drones, where a fleet of multiple trucks and multiple drones is routed.

\section*{Acknowledgment}

This material is based upon work supported by the National Research Foundation of Korea (NRF) grant funded by the Korean government (MSIT) [Grant No. RS-2023-00259550] and by the National Science Foundation [Grant No. 2032458].

\bibliographystyle{ormsv080-ck}
\bibliography{references}

\appendix

\begin{landscape}

	\section*{Appendix: Tables representing the details of experiments}

	\begin{table}[!htbp]
    
    \caption{Evaluation of HGA-TAC against optimal solutions for randomly generated instances}
    \begin{adjustbox}{max width=\linewidth} 
\begin{tabular}{clrrrrrrrrrrrrrrrrrrrrr}
            \toprule
            \multicolumn{1}{l}{}  &      & \multicolumn{7}{c}{$\alpha = 1$}                                                                                                                                                             & \multicolumn{7}{c}{$\alpha = 2$}                                                                                                                                                             & \multicolumn{7}{c}{$\alpha = 3$}                                                                                                                                                             \\
            \cmidrule(lr){3-9} \cmidrule(lr){10-16}  \cmidrule(lr){17-23}
            \multicolumn{1}{l}{}  &      & \multicolumn{2}{c}{MILP}                           & \multicolumn{5}{c}{HGA-TAC}                                                                                                          & \multicolumn{2}{c}{MILP}                           & \multicolumn{5}{c}{HGA-TAC}                                                                                                          & \multicolumn{2}{c}{MILP}                           & \multicolumn{5}{c}{HGA-TAC}                                                                                                          \\
\bottomrule
            \end{tabular}
\end{adjustbox}
\label{tab:exactGap}
\end{table}
\end{landscape}

\begin{landscape}	
		\begin{table}[!htbp]
			\centering
			\caption{(TSPD Set A$_u$, $\alpha=1$) Results of HGA-TAC and HGA-TAC$^+$ for \citet{agatz2018optimization} instances with $\alpha=1$. Times are in seconds, and Time* is as reported in  \citet{de2020variable}.}
			\label{tab:Agatz1Alpha1}
			\begin{adjustbox}{max width=\linewidth}        
			\begin{tabular}{ccrrrrrrrrrrrrrrrrr}
				\toprule
											   & \multicolumn{1}{c}{}  & \multicolumn{2}{c}{HGVNS}                          & \multicolumn{1}{c}{} & \multicolumn{2}{c}{DPS/25}                            & \multicolumn{1}{c}{} & \multicolumn{2}{c}{HGA-TAC}                           & \multicolumn{1}{c}{} & \multicolumn{2}{c}{Gap Over (\%)}                        & \multicolumn{1}{c}{} & \multicolumn{2}{c}{HGA-TAC$^+$}                         & \multicolumn{1}{c}{} & \multicolumn{2}{c}{Gap Over (\%)}                        \\
											   \cmidrule{3-4}  \cmidrule{6-7}  \cmidrule{9-13} \cmidrule{15-19}
				Dist                           & \multicolumn{1}{c}{$N$} & \multicolumn{1}{c}{Obj} & \multicolumn{1}{c}{Time*} & \multicolumn{1}{c}{} & \multicolumn{1}{c}{Obj} & \multicolumn{1}{c}{Time} & \multicolumn{1}{c}{} & \multicolumn{1}{c}{Avg.} & \multicolumn{1}{c}{Time} & \multicolumn{1}{c}{} & \multicolumn{1}{c}{HGVNS} & \multicolumn{1}{c}{DPS/25} & \multicolumn{1}{c}{} & \multicolumn{1}{c}{Avg.} & \multicolumn{1}{c}{Time} & \multicolumn{1}{c}{} & \multicolumn{1}{c}{HGVNS} & \multicolumn{1}{c}{DPS/25} \\
				\midrule
				\multirow{7}{*}{\rotatebox[origin=c]{90}{Uniform}}       & 10      & 289.82       & 0.14       &  & 265.20       & 0.01     &  & 265.20       & 0.23      &  & $-$8.49         & 0.00         &  & 265.20       & 0.70        &  & $-$8.49         & 0.00         \\
								   & 20      & 368.54       & 0.11       &  & 325.23       & 0.06     &  & 326.40       & 0.49      &  & $-$11.43        & 0.36         &  & 324.76       & 1.78        &  & $-$11.88        & $-$0.15        \\
								   & 50      & 559.20       & 3.71       &  & 498.20       & 0.26     &  & 498.45       & 1.55      &  & $-$10.86        & 0.05         &  & 493.84       & 6.54        &  & $-$11.69        & $-$0.87        \\
\bottomrule
			\end{tabular}
			\end{adjustbox}
		\end{table}
	\end{landscape}

	\begin{landscape}
		\begin{table}[]
			\centering
			\caption{(TSPD Set A$_u$, $\alpha=2$) Results of HGA-TAC and HGA-TAC$^+$ for \citet{agatz2018optimization} instances with $\alpha=2$. Times are in seconds, and Time* is as reported in  \citet{de2020variable}.}
			\label{tab:Agatz1Alpha2}
			\begin{adjustbox}{max width=\linewidth}
			\begin{tabular}{ccrrrrrrrrrrrrrrrrr}
				\toprule
				& \multicolumn{1}{c}{}  & \multicolumn{2}{c}{HGVNS}                          & \multicolumn{1}{c}{} & \multicolumn{2}{c}{DPS/25}                            & \multicolumn{1}{c}{} & \multicolumn{2}{c}{HGA-TAC}                           & \multicolumn{1}{c}{} & \multicolumn{2}{c}{Gap Over (\%)}                        & \multicolumn{1}{c}{} & \multicolumn{2}{c}{HGA-TAC$^+$}                         & \multicolumn{1}{c}{} & \multicolumn{2}{c}{Gap Over (\%)}                        \\
											   \cmidrule{3-4}  \cmidrule{6-7}  \cmidrule{9-13} \cmidrule{15-19}
	Dist                           & \multicolumn{1}{c}{$N$} & \multicolumn{1}{c}{Obj} & \multicolumn{1}{c}{Time*} & \multicolumn{1}{c}{} & \multicolumn{1}{c}{Obj} & \multicolumn{1}{c}{Time} & \multicolumn{1}{c}{} & \multicolumn{1}{c}{Avg.} & \multicolumn{1}{c}{Time} & \multicolumn{1}{c}{} & \multicolumn{1}{c}{HGVNS} & \multicolumn{1}{c}{DPS/25} & \multicolumn{1}{c}{} & \multicolumn{1}{c}{Avg.} & \multicolumn{1}{c}{Time} & \multicolumn{1}{c}{} & \multicolumn{1}{c}{HGVNS} & \multicolumn{1}{c}{DPS/25} \\
				\midrule
				\multirow{7}{*}{\rotatebox[origin=c]{90}{Uniform}}      & 10      & 233.20       & 0.13       &  & 218.57       & 0.01     &  & 218.04       & 0.17      &  & $-$6.50         & $-$0.24        &  & 218.08       & 0.54        &  & $-$6.49         & $-$0.22        \\
				& 20      & 293.60       & 0.85       &  & 277.51       & 0.11     &  & 274.87       & 0.36      &  & $-$6.38         & $-$0.95        &  & 274.19       & 1.25        &  & $-$6.61         & $-$1.20        \\
				& 50      & 420.80       & 2.30       &  & 420.60       & 0.59     &  & 415.69       & 1.38      &  & $-$1.21         & $-$1.17        &  & 408.00       & 4.96        &  & $-$3.04         & $-$2.99        \\
\bottomrule
			\end{tabular}
				\end{adjustbox}
		\end{table}
	\end{landscape}
	
	\begin{landscape}
		\begin{table}[]
			\centering
			\caption{(TSPD Set A$_u$, $\alpha=3$) Results of HGA-TAC and HGA-TAC$^+$ for \citet{agatz2018optimization} instances with $\alpha=3$. Times are in seconds, and Time* is as reported in  \citet{de2020variable}.}
			\label{tab:Agatz1Alpha3}
			\begin{adjustbox}{max width=\linewidth}
			\begin{tabular}{ccrrrrrrrrrrrrrrrrr}
				\toprule
				& \multicolumn{1}{c}{}  & \multicolumn{2}{c}{HGVNS}                          & \multicolumn{1}{c}{} & \multicolumn{2}{c}{DPS/25}                            & \multicolumn{1}{c}{} & \multicolumn{2}{c}{HGA-TAC}                           & \multicolumn{1}{c}{} & \multicolumn{2}{c}{Gap Over (\%)}                        & \multicolumn{1}{c}{} & \multicolumn{2}{c}{HGA-TAC$^+$}                         & \multicolumn{1}{c}{} & \multicolumn{2}{c}{Gap Over (\%)}                        \\
											   \cmidrule{3-4}  \cmidrule{6-7}  \cmidrule{9-13} \cmidrule{15-19}
	Dist                           & \multicolumn{1}{c}{$N$} & \multicolumn{1}{c}{Obj} & \multicolumn{1}{c}{Time*} & \multicolumn{1}{c}{} & \multicolumn{1}{c}{Obj} & \multicolumn{1}{c}{Time} & \multicolumn{1}{c}{} & \multicolumn{1}{c}{Avg.} & \multicolumn{1}{c}{Time} & \multicolumn{1}{c}{} & \multicolumn{1}{c}{HGVNS} & \multicolumn{1}{c}{DPS/25} & \multicolumn{1}{c}{} & \multicolumn{1}{c}{Avg.} & \multicolumn{1}{c}{Time} & \multicolumn{1}{c}{} & \multicolumn{1}{c}{HGVNS} & \multicolumn{1}{c}{DPS/25} \\
				\midrule
				\multirow{7}{*}{\rotatebox[origin=c]{90}{uniform}}       & 10      & 215.88       & 0.13       &  & 203.32       & 0.01     &  & 198.70       & 0.15      &  & $-$7.96         & $-$2.28        &  & 198.97       & 0.51        &  & $-$7.83         & $-$2.14        \\
				& 20      & 274.20       & 1.15       &  & 256.18       & 0.14     &  & 252.45       & 0.35      &  & $-$7.93         & $-$1.46        &  & 252.63       & 1.11        &  & $-$7.87         & $-$1.39        \\
				& 50      & 389.80       & 2.19       &  & 393.12       & 0.73     &  & 385.36       & 1.37      &  & $-$1.14         & $-$1.98        &  & 375.70       & 4.48        &  & $-$3.62         & $-$4.43        \\
\bottomrule
			\end{tabular}
		\end{adjustbox}
		\end{table}
	\end{landscape}
	
	\begin{landscape}
		\begin{table}
			\centering
		\caption{(TSPD Set A$_l$, limited) Results of HGA-TAC and HGA-TAC$^+$ for \citet{agatz2018optimization}'s instances with limited flying ranges, where $r$ is a parameter in the dataset that controls the maximum flying range. Times are in seconds.}
		\label{tab:Agatz2}
		\begin{adjustbox}{max totalheight=\textheight}
		\begin{tabular}{ccrrrrrrrrrrrrrrrrr}
						\toprule
						&         & \multicolumn{2}{c}{TSP-ep-all} &  & \multicolumn{2}{c}{DPS/25} &  & \multicolumn{2}{c}{HGA-TAC} &  & \multicolumn{2}{c}{Gap over (\%)} &  & \multicolumn{2}{c}{HGA-TAC$^+$} &  & \multicolumn{2}{c}{Gap over (\%)} \\
											   \cmidrule{3-4}  \cmidrule{6-7}  \cmidrule{9-13} \cmidrule{15-19}
			$N$                 & $r$       & Obj           & Time          &  & Obj         & Time      &  & Avg.         & Time      &  & TSP-ep-all      & DPS/25        &  & Avg.         & Time        &  & TSP-ep-all      & DPS/25        \\
			\midrule
			\multirow{6}{*}{10}  & 20      & 303.74                  & 0.01                       &                      & 303.74                  & 0.01                     &                      & 303.74               & 0.27              &  & 0.00            & 0.00       &  & 303.74               & 0.79                &  & 0.00            & 0.00       \\
			& 40      & 295.73                  & 0.01                       &                      & 295.73                  & 0.01                     &                      & 295.60               & 0.28              &  & $-$0.04           & $-$0.04      &  & 295.58               & 0.77                &  & $-$0.05           & $-$0.05      \\
			& 60      & 277.12                  & 0.01                       &                      & 277.12                  & 0.01                     &                      & 276.09               & 0.24              &  & $-$0.37           & $-$0.37      &  & 275.90               & 0.73                &  & $-$0.44           & $-$0.44      \\
			& 100     & 236.02                  & 0.01                       &                      & 236.02                  & 0.01                     &                      & 235.47               & 0.23              &  & $-$0.23           & $-$0.23      &  & 235.30               & 0.69                &  & $-$0.31           & $-$0.31      \\
\bottomrule
			\end{tabular}
			\end{adjustbox}
		\end{table}
	\end{landscape}

	\begin{landscape}
		\begin{table}[]
			\centering
			\caption{(TSPD Set B) TSPD results on the datasets of \citet{bogyrbayeva2023deep}. 
		Average cost and time values are reported on 100 problem instances for each size $N$. Times are in seconds, and Time* is as reported in \citet{bogyrbayeva2023deep}.
		Gap values are measured, in $\%$, over the Obj values of TSP-ep-all, and the Avg. values are used for HGA-TAC and HGA-TAC$^+$.
		}
			\label{tab:Aigerim}
			\begin{adjustbox}{max width=\linewidth}
			\begin{tabular}{cc rr r rrr r rrr r rrrr r rrrr}
			\toprule
								&         & \multicolumn{2}{c}{TSP-ep-all}                           & \multicolumn{1}{c}{} & \multicolumn{3}{c}{DPS/25}                                                                 & \multicolumn{1}{c}{} & \multicolumn{3}{c}{HM (4800)}                                                                  & \multicolumn{1}{c}{} & \multicolumn{4}{c}{HGA-TAC}                                                             & \multicolumn{1}{c}{} & \multicolumn{4}{c}{HGA-TAC$^+$}                                                            \\
								\cmidrule(lr){3-4} 
								\cmidrule(lr){6-8} 
								\cmidrule(lr){10-12}
								\cmidrule(lr){14-17}  
								\cmidrule(lr){19-22}

	Dataset                     & $N$       & \multicolumn{1}{c}{Obj} & \multicolumn{1}{c}{Time} & \multicolumn{1}{c}{} & \multicolumn{1}{c}{Obj} & \multicolumn{1}{c}{Time} & \multicolumn{1}{c}{Gap} & \multicolumn{1}{c}{} & \multicolumn{1}{c}{Obj} & \multicolumn{1}{c}{Time*} & \multicolumn{1}{c}{Gap} & \multicolumn{1}{c}{} & \multicolumn{1}{c}{Best}& \multicolumn{1}{c}{Avg.} & \multicolumn{1}{c}{Time} & \multicolumn{1}{c}{Gap} & \multicolumn{1}{c}{} & \multicolumn{1}{c}{Best} & \multicolumn{1}{c}{Avg.} & \multicolumn{1}{c}{Time} & \multicolumn{1}{c}{Gap} \\
	
	\midrule
	
	\multirow{4}{*}{\rotatebox[origin=c]{90}{\small Random}}     & 20      &        281.85                 &              0.10                   &                      &        281.85                &       0.01                       &   0.00                         &                      &          281.53               &             0.40                   &         $-$0.11                  &                   &      278.25              &          280.42            &           0.36          &     $-$0.51                &                         &            277.82                    &     279.54     &   1.19 &       $-$0.82             \\
\bottomrule                    
	\end{tabular}        	
	\end{adjustbox}
		\end{table}
	\end{landscape}
	
	\footnotesize

	\begin{longtable}{ccrrrrrrrrrrr}
		\label{tab:Murray}\\
		\caption{(FSTSP Set M) Results for FSTSP instances from \citet{murray2015flying}. MC and HGA20 represent the results reported by \citet{murray2015flying} and \citet{ha2020hybrid}, respectively. Times are in seconds. Gap values are measured, in $\%$, between the Avg. values of HGA-TAC and HGA-TAC$^+$ and the Avg. values of HGA20.}\\
		\toprule
	   \multicolumn{2}{c}{}               & \multicolumn{1}{c}{}                                                          & \multicolumn{2}{c}{HGA20}     & \multicolumn{4}{c}{HGA-TAC}     & \multicolumn{4}{c}{HGA-TAC$^+$}                                                                                                                                                                             \\
	   \cmidrule(lr){4-5}  \cmidrule(lr){6-9} \cmidrule(lr){10-13}
	Instance & $\epsilon$ & \multicolumn{1}{c}{MC} & \multicolumn{1}{c}{Best} & \multicolumn{1}{c}{Avg.}  & \multicolumn{1}{c}{Best} & \multicolumn{1}{c}{Avg.} & \multicolumn{1}{c}{Time} & \multicolumn{1}{c}{Gap} & \multicolumn{1}{c}{Best} & \multicolumn{1}{c}{Avg.} & \multicolumn{1}{c}{Time} & \multicolumn{1}{c}{Gap} \\
		\midrule
		
		\endfirsthead
		\caption{(Continued) }\\
	\toprule
	\multicolumn{2}{c}{}               & \multicolumn{1}{c}{}                                                          & \multicolumn{2}{c}{HGA20}     & \multicolumn{4}{c}{HGA-TAC}     & \multicolumn{4}{c}{HGA-TAC$^+$}                                                                                                                                                                             \\
	\cmidrule(lr){4-5}  \cmidrule(lr){6-9} \cmidrule(lr){10-13}
	Instance & $\epsilon$ & \multicolumn{1}{c}{MC} & \multicolumn{1}{c}{Best} & \multicolumn{1}{c}{Avg.}  & \multicolumn{1}{c}{Best} & \multicolumn{1}{c}{Avg.} & \multicolumn{1}{c}{Time} & \multicolumn{1}{c}{Gap} & \multicolumn{1}{c}{Best} & \multicolumn{1}{c}{Avg.} & \multicolumn{1}{c}{Time} & \multicolumn{1}{c}{Gap} \\
		\midrule
		
		\endhead

		437v1  & 20      & 56.468 & 56.468 & 56.468 & 56.393 & 56.393 & 0.38 & $-$0.13\% & 56.393 & 56.393 & 1.16 & $-$0.13\% \\
		437v1  & 40      & 50.573 & 50.573 & 50.573 & 50.573 & 50.573 & 0.39 & 0.00\%  & 50.573 & 50.573 & 1.18 & 0.00\%  \\
		437v2  & 20      & 53.207 & 53.207 & 53.207 & 53.207 & 53.207 & 0.35 & 0.00\%  & 53.207 & 53.207 & 1.11 & 0.00\%  \\
		437v2  & 40      & 47.311 & 47.311 & 47.311 & 47.311 & 47.311 & 0.37 & 0.00\%  & 47.311 & 47.311 & 1.13 & 0.00\%  \\
		437v3  & 20      & 53.687 & 53.687 & 53.687 & 54.664 & 54.664 & 0.33 & 1.82\%  & 54.664 & 54.664 & 1.11 & 1.82\%  \\
		437v3  & 40      & 53.687 & 53.687 & 53.687 & 53.687 & 53.687 & 0.39 & 0.00\%  & 53.687 & 53.687 & 1.16 & 0.00\%  \\
		437v4  & 20      & 67.464 & 67.464 & 67.464 & 67.464 & 67.464 & 0.34 & 0.00\%  & 67.464 & 67.464 & 1.10 & 0.00\%  \\
		437v4  & 40      & 66.487 & 66.487 & 66.487 & 66.487 & 66.487 & 0.39 & 0.00\%  & 66.487 & 66.487 & 1.17 & 0.00\%  \\
\bottomrule
	\end{longtable}

	\newpage

		\begin{longtable}[H]{lrrrrrrrrrrrrr}
			\label{tab:Ha}\\
			\caption{(FSTSP Set H) Results for FSTSP instances from \citet{ha2018min}. Times are in seconds, and Time* represents the time reported in \citet{ha2018min}. Gap is measured between the Avg. values of HGA-TAC and HGA-TAC$^+$ and the Avg. values of HGA20. }\\
			\toprule
					 & \multicolumn{3}{c}{HGA20}                                                        & \multicolumn{1}{c}{} & \multicolumn{4}{c}{HGA-TAC}                                                                                 & \multicolumn{1}{c}{} & \multicolumn{4}{c}{HGA-TAC$^+$}                                                                               \\
					 \cmidrule{2-4}   \cmidrule{6-9}   \cmidrule{11-14}
			Instance & \multicolumn{1}{c}{Best} & \multicolumn{1}{c}{Avg.} & \multicolumn{1}{c}{Time*} & \multicolumn{1}{c}{} & \multicolumn{1}{c}{Best} & \multicolumn{1}{c}{Avg.} & \multicolumn{1}{c}{Gap} & \multicolumn{1}{c}{Time} & \multicolumn{1}{c}{} & \multicolumn{1}{c}{Best} & \multicolumn{1}{c}{Avg.} & \multicolumn{1}{c}{Gap} & \multicolumn{1}{c}{Time} \\
			\midrule
			\endfirsthead
			\caption{(Continued)}\\
			\toprule
					 & \multicolumn{3}{c}{HGA20}                                                        & \multicolumn{1}{c}{} & \multicolumn{4}{c}{HGA-TAC}                                                                                 & \multicolumn{1}{c}{} & \multicolumn{4}{c}{HGA-TAC$^+$}                                                                               \\
					 \cmidrule{2-4}   \cmidrule{6-9}   \cmidrule{11-14}
			Instance & \multicolumn{1}{c}{Best} & \multicolumn{1}{c}{Avg.} & \multicolumn{1}{c}{Time*} & \multicolumn{1}{c}{} & \multicolumn{1}{c}{Best} & \multicolumn{1}{c}{Avg.} & \multicolumn{1}{c}{Gap} & \multicolumn{1}{c}{Time} & \multicolumn{1}{c}{} & \multicolumn{1}{c}{Best} & \multicolumn{1}{c}{Avg.} & \multicolumn{1}{c}{Gap} & \multicolumn{1}{c}{Time} \\
			\midrule
			\endhead
			B1      & 115.65 & 116.43 & 45.60  &  & 116.92 & 117.23 & 0.69  & 2.73  &  & 115.11 & 115.89 & $-$0.46 & 12.26 \\
	B2      & 118.39 & 118.39 & 19.80  &  & 113.93 & 114.17 & $-$3.56 & 1.69  &  & 113.50 & 113.82 & $-$3.86 & 9.06  \\
	B3      & 116.21 & 116.39 & 34.20  &  & 114.09 & 114.75 & $-$1.41 & 2.59  &  & 112.25 & 113.52 & $-$2.46 & 10.57 \\
	B4      & 118.71 & 119.26 & 28.20  &  & 119.85 & 121.77 & 2.11  & 2.05  &  & 119.19 & 120.88 & 1.35  & 10.38 \\
	B5      & 115.78 & 115.91 & 34.80  &  & 113.77 & 114.38 & $-$1.32 & 1.84  &  & 112.77 & 114.44 & $-$1.27 & 6.02  \\
	B6      & 114.31 & 115.46 & 52.80  &  & 114.04 & 115.25 & $-$0.18 & 2.87  &  & 112.11 & 112.69 & $-$2.40 & 12.07 \\
	B7      & 115.52 & 115.63 & 37.20  &  & 115.53 & 117.84 & 1.91  & 2.67  &  & 113.63 & 115.02 & $-$0.52 & 12.56 \\
	B8      & 117.90 & 118.04 & 46.80  &  & 113.25 & 114.13 & $-$3.32 & 2.41  &  & 113.25 & 113.70 & $-$3.68 & 10.00 \\
\bottomrule
	\end{longtable}

\end{document}